%% file: root.tex
\documentclass[lettersize,journal]{IEEEtran}  % Comment this line out if you need a4paper
\input{packages.tex}
\usepackage{setspace}
\usepackage[colorlinks,citecolor=red,urlcolor=blue,bookmarks=false,hypertexnames=true]{hyperref} 

\renewcommand{\baselinestretch}{.986}

\def\argmin{\mathop{\arg\min}\limits}
\def\argmax{\mathop{\arg\max}\limits}

\newcommand{\crl}[1]{\left\{#1\right\}}

\makeatletter
\newcommand{\customlabel}[2]{%
   \protected@write \@auxout {}{\string \newlabel {#1}{{#2}{\thepage}{#2}{#1}{}} }%
   \hypertarget{#1}{#2}
}
\makeatother

\title{\LARGE \bf
Distributed Variational Inference for Online Supervised Learning %and Filtering
}
\author{Parth Paritosh, Nikolay Atanasov and Sonia Mart\'{i}nez%
  \thanks{The authors are with the Contextual Robotics Institute,
    University of California San Diego, 9500 Gilman Dr, La Jolla, CA
    92093 {\tt\small \{pparitos,natanasov,soniamd\}@ucsd.edu}.
    We gratefully acknowledge support from ONR N00014-19-1-2471, 
    ARL DCIST CRA W911NF17-2-0181 and NSF FRR CAREER 2045945.}%
}

\begin{document}

\maketitle
\thispagestyle{empty}
\pagestyle{empty}

%%%%%%%%%%%%%%%%%%%%%%%%%%%%%%%%%%%%%%%%%%%%%%%%%%%%%%%%%%%%%%%%%%%%%%%%%%%%%%%%
\begin{abstract}
Developing efficient solutions for inference problems in intelligent sensor networks is crucial for the next generation of location, tracking, and mapping services. This paper develops a scalable distributed probabilistic inference algorithm that applies to continuous variables, intractable posteriors and large-scale real-time data in sensor networks. In a centralized setting, variational inference is a fundamental technique for performing approximate Bayesian estimation, in which an intractable posterior density is approximated with a parametric density. Our key contribution lies in the derivation of a separable lower bound on the centralized estimation objective, which enables distributed variational inference with one-hop communication in a sensor network. 
Our distributed evidence lower bound (DELBO) consists of a weighted sum of observation likelihood and divergence to prior densities, and its gap to the measurement evidence is due to consensus and modeling errors. 
To solve binary classification and regression problems while handling streaming data, we design an online distributed algorithm that maximizes DELBO, and specialize it to Gaussian variational densities with non-linear likelihoods.  The resulting distributed Gaussian variational inference (DGVI) efficiently inverts a $1$-rank correction to the covariance matrix. Finally, we derive a diagonalized version for online distributed inference in high-dimensional models, and apply it to multi-robot probabilistic mapping using indoor LiDAR data.
\end{abstract}

%  Our key contribution lies in the
%  derivation of a separable lower bound on the centralized estimation
%  objective, which enables distributed variational inference. 
%  Our distributed evidence lower bound (DELBO) consists of a weighted sum of likelihood and
%  divergence terms constituting the distributed estimation objective.
%  We show that the gap from the bound is due to residual-consensus and 
%  posterior-modeling terms. To solve binary classification
%  and regression problems while handling streaming data, we
%  design a real-time distributed algorithm that maximizes DELBO, and
%  specialize it to Gaussian estimates with non-linear likelihoods.  We
%  further modify the resulting distributed Gaussian variational
%  inference (DGVI) to minimize matrix inverse computations for
%  computational efficiency. Finally, we derive a diagonalized version
%  for real-time distributed inference in high-dimensional models, and
%  apply it to multi-robot probabilistic mapping using indoor LiDAR
%  datasets\footnote{Source code available at
%    \url{https://github.com/pptx/distributed-mapping}}.

% Yes, we do quantify the objective function gap. I think your question is valid in the sense that this information is never clearly quantified.

\input{section1_introduction.tex}

\input{section2_problemf.tex}

\input{section3_background.tex}

\input{section4_delbo.tex}
\input{section5_dgvi.tex}

\input{section6_results.tex}

%%%%%%%%%%%%%%%%%%%%%%%%%%%%%%%%%%%%%%%%%%%%%%%%%%%%%%%%%%%%%%%%%%%%%%%%%%%%%%%%%%%%%%%%%%
\section{Conclusion}
%%%%%%%%%%%%%%%%%%%%%%%%%%%%%%%%%%%%%%%%%%%%%%%%%%%%%%%%%%%%%%%%%%%%%%%%%%%%%%%%%%%%%%%%%%

Analogous to the evidence lower bound (ELBO) in variational inference, this paper derived a distributed evidence lower bound (DELBO) on the observation evidence in multi-agent estimation problems. Optimizing the components of DELBO separately at each agent led to a distributed variational inference algorithm. We derived a version of the algorithm with Gaussian variational distributions and applied it to multi-robot mapping problems using streaming range measurements. Our distributed VI algorithm handles general non-linear observation likelihood models efficiently making it a promising approach for network estimation problems with various machine learning models. A potential avenue for future work is to improve the communication efficiency of the algorithm by limiting the number of communication rounds and the number of actively communicating agents or by allowing agents to share subsets of their local parameter estimates.

%In this work, we compute a separable lower bound on the estimated posterior density 
%and named it DELBO analogous to standard variational inference.
%Then, we optimize the agent specific DELBO component to recover a distributed estimation algorithm. 
%Upon specializing this algorithm to Gaussian density to obtain a distributed Gaussian variational inference algorithm,
%which is then applied successfully to distributed mapping problem over multiple data sets.
%Improving the communication efficiency is a valuable future direction for this work.
%Such efficiency can also be achieved by designing performance indicators to limit the number of communication rounds,
% limiting the number of actively communicating agents and their shared parameters.

\bibliographystyle{abbrv}
\bibliography{graphprediction.bib}
\appendix

\input{appendix_simplification.tex}

%\input{lik_est.tex}
%\input{appendix.tex}
%%%%%%%%%%%%%%%%%%%%%%%%%%%%%%%%%%%%%%%%%%%%%%%%%%%%%%%%%%%%%%%%%%%%%%%%%%%%%%%%%%%%%%%%%%

\end{document}

%% file: packages.tex
\usepackage{amssymb}
\usepackage{amsthm}
\usepackage[fleqn]{amsmath}
\usepackage{epsfig}
\usepackage[noend]{algorithm2e}
\usepackage{amsfonts}
\usepackage{amsmath}
\usepackage{multirow}
\usepackage{pdflscape}
\usepackage{epstopdf}
\usepackage{listings}  
\usepackage{dsfont}
\usepackage[table,usenames,dvipsnames]{xcolor}      % color
\usepackage{tikz}

\newcommand{\Pe}{\mathbb{P}}

\newcommand{\expect}{\mathbb{E}}

\newcommand{\BlackBox}{\rule{1.5ex}{1.5ex}}  % end of proof
\ifdefined\proof
    \renewenvironment{proof}{\par\noindent{\bf Proof\ }}{\hfill\BlackBox\\[2mm]}
\else
    \newenvironment{proof}{\par\noindent{\bf Proof\ }}{\hfill\BlackBox\\[2mm]}
\fi

\newcommand{\calN}{\mathcal{N}}

\newcommand{\calF}{\mathcal{F}}

\newtheorem{theorem}{Theorem}
\newtheorem{corollary}{Corollary}
\newtheorem{lemma}{Lemma}

\newtheorem{assumption}{Assumption}
\newtheorem{problem}{Problem}
\theoremstyle{remark}
\newtheorem{remark}{Remark}
\newtheorem{example}{Example}

\newcommand{\real}{\ensuremath{\mathbb{R}}}

\newcommand{\nodes}{\mathcal{V}}
\newcommand{\edges}{\mathcal{E}}
\newcommand{\graph}{\mathcal{G}}

\newcommand{\diag}[1]{\operatorname{diag}(#1)}

\newcommand{\qz}[1]{\operatorname{\ell}_{#1}}
\newcommand{\vect}[1]{\boldsymbol{#1}}
\newcommand{\ones}[1]{\vect{1}_{#1}}

\newcommand{\neighbor}[1]{\mathcal{V}_{#1}}

\newcommand{\KL}{\operatorname{\mathrm{KL}}}

%% file: section1_introduction.tex
%%%%%%%%%%%%%%%%%%%%%%%%%%%%%%%%%%%%%%%%%%%%%%%%%%%%%%%%%%%%%%%%%%%%%%%%%%%%%%%%
\section{Introduction}
%%%%%%%%%%%%%%%%%%%%%%%%%%%%%%%%%%%%%%%%%%%%%%%%%%%%%%%%%%%%%%%%%%%%%%%%%%%%%%%%

%%%%%%%%%%%%%%%%%%%%%%%%%%%%%%%%%%%%%%%%%%%%%
% What is the utility of variational inference?
% What sort of problems have been solved? Which domains? What about robotics?
% Brief history of Variational inference
% Distributed algorithms
% Distributed variational inference
% Contributions
%%%%%%%%%%%%%%%%%%%%%%%%%%%%%%%%%%%%%%%%%%%%%%
%\margin{The intro is still a bit long, I'm trying to cut it down}

Modern cyber-physical networks composed of autonomous vehicles and IoT devices continuously generate large volumes of data.
Estimating variables and parameters of interest from the data efficiently and accurately subject to the computation, communication, and storage constraints of the network devices is a critical problem. Distributed estimation methods are an effective way to handle these constraints, while 
avoiding the single-point failures in centralized estimation techniques.
% avoiding the pitfall of a single-point of failure that arises in centralized estimation techniques.

Bayesian inference is a probabilistic estimation method that accumulates observation likelihood information to compute the (posterior) distribution of the variables of interest conditioned on the observations. This is especially useful in prediction problems because the uncertainty quantification provided by the posterior distribution helps limit overconfidence about the best estimate. Yet, the Bayesian approach comes at a cost, which is computational intractability for general observation models. This has given rise to approximate inference rules, including expectation propagation and variational inference, which can provide more efficient posterior computations. This work investigates the design of a distributed variational inference algorithm that can handle continuous variables, intractable posteriors, and large datasets in sensor networks.
% EP: \cite{TM:13}
% Fixed point for exponential distributions only. https://dotnet.github.io/infer/userguide/Working%20with%20different%20inference%20algorithms.html

\paragraph*{Contributions}
This paper derives a distributed version of the evidence lower bound (ELBO) \cite{SM-MH-DB:17} used in variational inference to enable posterior density approximation as an optimization problem over the space of probability density functions. Our distributed ELBO (DELBO) leads to an optimization objective that can be decomposed into separate local objectives for each node in a graph, enabling fully distributed inference with one-hop communication among the nodes. Focusing on Gaussian variational densities, we obtain explicit updates for nonlinear observation likelihoods in the form of a distributed Gaussian variational inference (DGVI) algorithm. Further specialization to diagonal Gaussian densities enables efficient large-scale inference. We apply these algorithms to achieve distributed probabilistic classification in multi-robot mapping problems using streaming LiDAR data.

\paragraph*{Related work}
%%%%%%%%%%%%%%%%%%%%%%%%%%%%%%%%%%%%%%%%%%%%%%%%%%%%%%%%%%%%%%%%%%%%%%%%%%%%%%%%
Variational inference (VI) \cite{PSL:86, MIJ-ZG-:99} is an approximation to standard Bayesian
inference that handles general observation models in state estimation \cite{SG-JP:17}, learning from demonstrations
\cite{TS-AG:20}, and simultaneous localization and mapping \cite{TDB-JRF-DJY:20}. 
VI has also been used to train autoencoders and deep generative models \cite{DJR-SM-DW:14, DPK-MW:19}. 
In VI \cite{TSJ-MIJ:00}, posterior probability density functions (pdfs) are calculated to maximize a lower 
bound (ELBO) on measurement evidence containing divergence to the true posterior pdf. 
See the early work \cite{ZG-MB:00}, which computes such updates for conjugate families of prior
and likelihood distributions.  However, many applications require
non-linear log-likelihood models and non-conjugate priors.  
Posterior sampling techniques relying on sequential or Hamiltonian Monte Carlo 
sampling \cite{HD-YZ-JL:13, VS-AQ:08} produce posterior approximations by collecting samples from a Markov chain model. 
Unfortunately, in high-dimensional problems, the number of samples required to obtain useful approximations is computationally prohibitive. 
Instead, stochastic optimization algorithms \cite{MH-DB-CW-JP:13} are applied to the ELBO objective 
to learn an approximate posterior density from noisy gradients. 
Under some assumptions, stochastic gradient descent can even be interpreted as a Markov chain
to infer posteriors~\cite{SM-MH-DB:17}. 
We rely on gradient descent to derive updates specialized to a class of parametric families for analytic computation.  

A popular adaptation of stochastic optimization in VI takes the form
of Gaussian variational inference (GVI), where a Gaussian posterior is
estimated for arbitrary data likelihoods. Barfoot et al. \cite{TDB-JRF-DJY:20} estimate blocks of the full
covariance matrix to develop an online GVI algorithm. 
However, none of these methods develop a distributed framework for inference. 
Distributed algorithms allow agents to share computational
load across the network, and avoid raw data transmission. 
Decentralized algorithms perform better in practice \cite{XL:17} as they reduce the load on the busiest node and avoid
single point failures. In what follows, we specialize our review in
probabilistic inference to distributed estimation and optimization and federated learning literature.

Federated learning was originally developed for learning models over
data repositories \cite{PK:21} in server-client architectures, such as edge computing. 
Federated averaging was shown to perform accurate inference on non-IID data distributions over this
architecture in \cite{BM:17}, with posterior density averaging in \cite{MA-JG-EX-AR:20}. 
There have been recent extensions to fully decentralized settings with non-IID data \cite{TDB-CVN-SS-RET:18, XZ-YL-WL:22, ST-DL-BW:22, SZ-GYL:23}. In Gaussian inference, the covariance matrix is updated from batches of data in federated
settings \cite{VO-DN-MS:18}. More recently, model aggregation has
been studied over arbitrary communication networks
\cite{XW-AL-TJ-FK:22}. Their work draws from the social learning analysis to upper bound the error in the estimated pdf 
but the updates rely on sample-intensive Monte Carlo methods.

In contrast, distributed estimation and optimization problems such as distributed least squares 
require consistent estimates for arbitrary connectivity. Their solutions result in algorithms minimizing a sum of separable objective functions subject to a consensus constraint; 
see the recent survey on distributed learning via parametric optimization~\cite{XC-TB:23}.
Variants of stochastic gradient descent are widely used to obtain 
consistent solutions with inexact gradient samples at agents, 
but most are limited to finite dimensional point estimates~\cite{YT:19}.
Additionally, their reliance on strongly convex objectives for guarantees renders 
them incompatible with the divergence terms in a VI inference objective.
In addition to addressing this issue, we aim to perform probabilistic inference 
in presence of noisy gradients using data streamed over a connected network. 
This differentiates from previous work \cite{JC:21, CAU-AO-AN:22, PP-NA-SM:22}, 
which presents a class of distributed Bayesian algorithms that estimate pdfs for localization problems.  In
particular, these implementations are restricted to conditionally conjugate families of distributions. 
To relax this assumption, we aim to combine VI methods with such 
distributed Bayesian algorithms with noisy gradients for arbitrarily connected networks. 
An existing VI algorithm \cite{JH-CL:15} solves a distributed inference problem similar to this work, but our solution avoids the reliance on computationally expensive sampling. We instead look at specific classification, regression, and filtering models to obtain analytical updates.

The rest of the manuscript is organized as follows. Section~\ref{sec:problem}
formulates the distributed inference problem over the space of pdfs. 
Section~\ref{sec:background} introduces variational inference and derives the ELBO. Section~\ref{sec:dgvi} devises a distributed version of the evidence lower bound which leads to distributed variational inference. Tractable iterative update rules are presented in section V for Gaussian family densities. These algorithms are demonstraed in multi-robot mapping problems in Section~\ref{sec:results}.

%% file: section2_problemf.tex
%%%%%%%%%%%%%%%%%%%%%%%%%%%%%%%%%%%%%%%%%%%%%%%%%%%%%%%%%%%%%%%%%%%%%%%%%%%%%%%%
\section{Problem formulation: Distributed inference}
\label{sec:problem}
%%%%%%%%%%%%%%%%%%%%%%%%%%%%%%%%%%%%%%%%%%%%%%%%%%%%%%%%%%%%%%%%%%%%%%%%%%%%%%%%

% the source and/or relative agent positions in localization problems, or an occupancy map in distributed mapping tasks. To solve any of these inference problems, 

% Three paragraphs
Consider $n$ agents $\nodes = \{1, \dots, n\}$ aiming to estimate an unknown variable $\theta \in \real^l$ cooperatively. The
variable $\theta$ may represent a measurement source in environmental monitoring, relative agent positions in a localization problem, or environment occupancy in a mapping problem. The agents need to
address two main challenges: 1) observations are received online and are noisy and 2) the observations are partially informative about $\theta$ due to the agents' states and limited sensing capabilities. Therefore, the agents need to cooperate to learn an accurate and consistent estimate of $\theta$. Suppose that agent $i$ receives observation $z_{i,t} \in \real^d$, at each time $t$, according to a known observation likelihood model $\qz{i}(z_{i,t}|\theta)$. We make the following assumption.

\begin{assumption}[Independence]
  \label{assume:independence}
  The observations $z_t = \{z_{i,t}\}_{i \in \nodes}$ received by the agent
  network at any time $t$ are independent samples of the likelihood
  $\qz{}(z_t|\theta) = \prod_{i \in \nodes} \qz{i}(z_{i,t}|\theta)$.
\end{assumption}

To account for stochastic and partially informative observations,
the agents are to cooperatively agree on a probability distribution
$p(\theta)$ over the variable $\theta$. This cooperation is
enabled by communication over a strongly connected digraph,
$\graph = (\nodes, \edges)$, with edge set $\edges \subseteq \nodes \times \nodes$.
The edge $(i,j) \in \edges$ implies that node~$j$ transmits information to node~$i$.
Recall that a graph is strongly connected \cite{FB-JC-SM:09} if there exist a
directed path between any two nodes in the network, thus allowing flow of information across nodes. 
The allowable information flow is captured using a non-negative, irreducible weighted adjacency matrix $A$, 
such that with $A_{ij}>0$ only if $(i,j) \in \edges$. 
Using the Sinkhorn's algorithm \cite{RS-PK:67},
the adjacency matrix can be made doubly stochastic, i.e.,
$A\ones{n} = A^{\top}\ones{n} = \ones{n}$, where $\ones{n}$ is a vector of ones. Therefore, we assume the following.

\begin{assumption}[Connectivity]
    \label{assume:connect}
    The weighted adjacency matrix $A$ representing the communication
    graph $\graph$ is doubly stochastic $A\ones{n} = A^{\top}\ones{n} = \ones{n}$ and strongly connected.
\end{assumption}
 
The collaborative network thus aims to estimate the density
$p(\theta|z_{\leq t})$ at time~$t$, where $z_{\leq t}$ represents
observations collected by all agents until time $t$. 
We assume that the selected agent priors $p_{i,0}(\theta)$ are positive over the feasible domain in $\theta$.
Based on this, we state the problem formally next.
% Informative prior assumption: We assume that the priors at agents place non-zero weight on the feasible domain. 
\begin{problem}
Given observations $\{z_{i,t}\}$ sampled from the agent observation models $\qz{i}(z_{i,t}|\theta)$, and priors $\crl{p_{i,0}(\theta)}$ over an unknown parameter $\theta$, compute a posterior pdf $p_i(\theta|z_{\leq t})\in \calF$, where $\calF$ is a known pdf family and subject to consensus constraint $p_{i}(\theta|z_{\leq t}) = p_j(\theta|z_{\leq t})$, for $i,j \in \nodes$ and any $t\ge 0$.  
\end{problem}
% \NA{Shouldn't the left-hand side depend on $t$ and the right-hand side on $i$?}
% \subset \mathcal{P}(\real^l)

%  Given data sequences $\{z_{i,t}\}$ sampled from 
%  an unknown pdfs $p_i^\star(\theta)$ 
%  with arbitrary likelihood $\qz{i}(z_{i,t}|\theta)$, 
%  find an online approximation $p_i(\theta) \approx p(\theta|z_{\leq t})$
%  for all $i \in \nodes$, subject to a model constraint $p_i(\theta) \in \calF$, 
%  and a consensus constraint $p_i(\theta) = p(\theta)$, for all $i \in \nodes$, 
%  known family $\calF$ and any $t\ge 0$.   

%% file: section3_background.tex
%%%%%%%%%%%%%%%%%%%%%%%%%%%%%%%%%%%%%%%%%%%%%%%%%%%%%%%%%%%%%%%%%%%%%%%%%%%%%%%%%%%%%%%%%
\section{Background}
\label{sec:background}
%%%%%%%%%%%%%%%%%%%%%%%%%%%%%%%%%%%%%%%%%%%%%%%%%%%%%%%%%%%%%%%%%%%%%%%%%%%%%%%%%%%%%%%%%%
% We focus on a single agent, 
In this section, we review the centralized variational inference (VI) approach, 
that we later connect to the proposed distributed VI setting.
The classic Bayes approach calculates the posterior distribution of a
parameter $\theta$ at time $t$ as,
\begin{equation} \label{eqn:Bayes}
  p(\theta| z_{\leq t}) %= \frac{p(z_t|\theta)p(z_{<t}|\theta)p(\theta)}{p(z_{\leq t})}
   = \frac{\qz{}(z_t|\theta)p(\theta|z_{<t})}{p(z_t|z_{< t})}, 
\end{equation}
by which the posterior $p(\theta| z_{\leq t})$ is proportional to
the likelihood $\qz{}(z_{t}|\theta)$ and the prior $p(\theta| z_{< t})$. When the prior is conditionally
conjugate to the likelihood, it is well known that an analytical computation of \eqref{eqn:Bayes} is feasible \cite{DF:97}. 
%, such as the class of exponential families.
For instance, a Gaussian prior with Gaussian linear likelihood density functions
 leads to the standard Gaussian posterior update. 
% otherwise known as the Kalman filter.
Yet, the exact calculation of \eqref{eqn:Bayes} for general prior-likelihood pairs is not possible, as the computation of the normalization factor $p(z_t|z_{< t}) = \int \qz{}(z_t|\theta)p(\theta|z_{<t}) d\theta$ is intractable.

The Bayesian inference rule \eqref{eqn:Bayes} can be
obtained as the solution to a maximization problem over the space $\mathcal{P}(\real^l)$ of
probability distributions $q(\theta)$ on $\theta \in \real^l$. This maximization is performed over the
so-called Evidence Lower Bound (ELBO). The VI approach specializes
this problem to a finite-dimensional family of pdfs, $\calF \subset \mathcal{P}(\real^l)$, which often includes exponential
densities \cite{CZ-JB-HK:18}.
% Instead of learning the
% accurate but intractable posterior, VI algorithms approximate the
% posterior $p(\theta|z_{\leq t})$ with the pdf $q_t(\theta) \in \calF$.
%be given as maximizing a lower bound
%to the normalizing factor in the Bayesian update
%rule~\eqref{eqn:Bayes}.
% For completeness and clarity, we briefly reproduce it here to highlight the parallel
% with the proposed distributed version.
Despite ELBO's ubiquity in the VI literature, we briefly reproduce it
here for the sake of completeness and clarify the
parallel with the proposed distributed version.
To proceed, for pdfs $p, q \in \calF$, we define the differential entropy
$H(q(\theta)) = - \expect_{q(\theta)} \log q(\theta)$
and KL-divergence $\KL [q(\theta) || p(\theta)] 
= \expect_{q(\theta)} \left[ \log \frac{q(\theta)}{p(\theta)}\right]$.
% The knowledge of the posterior follows from the joint distribution 
% $p(\theta,z_{\leq t})/\int p(\theta,z_{\leq t}) dz_{\leq t}$,
% whose intractability arises from the normalization factor $\int p(\theta,z_{\leq t}) dz_{\leq t}$.
% Therefore, the posterior probability is best approximated when the normalization 
% factor $p(z_t|z_{< t})$ is minimized. 
 % Introduce as an explanation of VI, q_t lives in \calF.

\begin{lemma}\label{lemma:elbo}
  Given a pdf $q(\theta)$, the normalization factor $p(z_t|z_{< t})$ in~\eqref{eqn:Bayes} 
  is lower bounded by the ELBO, %Evidence Lower Bound (ELBO),
\begin{align*}
  \underset{q(\theta)}{\expect} [\log \qz{}(z_t|\theta)- \log (q(\theta)) 
  + \log p(\theta| z_{< t})].  
\end{align*}
\end{lemma}

\begin{proof}
  Using \eqref{eqn:Bayes}, the normalization factor is expressed in terms of the
  approximated posterior pdfs as,
  \begin{align}
    & \log p(z_t|z_{< t}) = \underset{q(\theta)}{\expect} 
      \left[ \log \frac{\qz{}(z_t|\theta) p(\theta|z_{<t}) q(\theta)}{p(\theta| z_{\leq t}) q(\theta)} \right] \nonumber \\
    & = \underset{q(\theta)}{\expect} [\log \qz{}(z_t|\theta)] - \KL [q(\theta) || p(\theta| z_{< t})] \nonumber \\
    & \quad + \KL [q(\theta) || p(\theta| z_{\leq t})] \nonumber \\
    & \geq \underset{q(\theta)}{\expect} [\log \qz{}(z_t|\theta)] - \KL [q(\theta) || p(\theta| z_{< t})] \nonumber \\
    & = \underset{q(\theta)}{\expect} [\log \qz{}(z_t|\theta)
      -\log q(\theta) + \log p(\theta| z_{< t})] . \label{eqn:ELBO}% \\
    % & \approx \underset{q_t(\theta)}{\expect} [\log \qz{}(z_t|\theta)] + H(q_t(\theta)) 
    % + \underset{q_t(\theta)}{\expect} [\log q_{t-1}(\theta)] \nonumber 
  \end{align}  
  Since the argument $\theta$ in $q$ is independent of the data
  $z_{\leq t}$, the expectation does not alter the value of the
  log-normalization.  The non-negative variational gap term in the
  second line, $\KL [q(\theta) || p(\theta| z_{\leq t})]$, is
  discarded to obtain the ELBO.  
\end{proof}

To continue iteratively in VI, we find the best approximating pdf $q_t(\theta)$ 
of the posterior $p(\theta|z_{\leq t})$ in a family $\calF$ for each time $t$.
The previous posterior $p(\theta| z_{< t})$ in the ELBO term is replaced with the known
$q_{t-1}(\theta)$ and the next posterior $q_t(\theta)$ is chosen to maximize the ELBO,
\begin{align}
  \label{eqn:c_geom_update}
  q_{t}(\theta) & \in \argmin_{q(\theta) \in \calF} \left\{ - \langle q, \log \qz{}(z_t|\theta) \rangle 
  + \KL[q||q_{t-1}]\right\} 
\end{align}
% The approximation $q_t(\theta)$ is selected within the family $\calF$
% to minimize its divergence with the true posterior.
% {\color{magenta} Variational parameters}
% \margin{what does variational parameters mean? Does this mean we
%   restrict ourselves to a finite-dimensional parameteric family
%   $\calF$? I prefer we said it that way.}  
%\NA{This is confusing. Is $\theta$ the parameter of $q_t$ or is it the observation generating parameter?}
When the pdf $q_t(\theta)$ is parametrized, the problem aims to find its hyperparameters minimizing the divergence to the true posterior. 
The lower bound explains the modeling error induced by the choice of the distributional family $\calF$.
VI also admits the interpretation of finding the best $\calF$-constrained optimization solution to a minimization objective; see~\cite[Section 2.2]{JK-JJ-TD:22} for more information.   

%\margin{I've read up to here}

%% file: section4_delbo.tex
%%%%%%%%%%%%%%%%%%%%%%%%%%%%%%%%%%%%%%%%%%%%%%%%%%%%%%%%%%%%%%%%%%%%%%%%%%%%%%%%%%%%%
\section{Distributed evidence lower bound}
%%%%%%%%%%%%%%%%%%%%%%%%%%%%%%%%%%%%%%%%%%%%%%%%%%%%%%%%%%%%%%%%%%%%%%%%%%%%%%%%%%%%%

In this section, we derive a distributed version of the VI optimization problem in Eqn.~\ref{eqn:c_geom_update}.
In this setting, the $n$ agents follow Assumption~\ref{assume:independence} to collect data independently. Each agent $i$ maintains its own local pdf $p_i(\theta | z_{<t})$ estimating the centralized density $p(\theta | z_{<t})$ over the parameter $\theta$ at time $t$. 
Since the agents have their own likelihood models, their estimated densities may not be equal.
Using the geometric average of the local pdfs $p(\theta | z_{<t}) 
\propto \prod_{i=1}^n p_i(\theta | z_{<t})^{1/n}$ to represent the centralized prior, we can rewrite Bayes' rule as,
\begin{align}
  \label{eqn:dBayes}
  p(\theta| z_{\leq t}) = \frac{\prod_{i \in \nodes} \qz{i}(z_{i,t} | \theta) p_i(\theta|z_{<t})^{1/n}}{p(z_t | z_{<t})}.
\end{align}
As before, we start by computing a lower bound on the normalization term 
analogous to the ELBO in~\eqref{eqn:ELBO}. To obtain a separable version of the VI objective, the agent likelihoods and priors are separated in the lower bound. Maximizing the separable components at each agent yields a distributed probabilistic inference algorithm, where each component contains the corresponding agent's private observations. 
\begin{theorem}\label{thm:delbo}
  Given agent pdfs $q_{i,t}(\theta) = q_{t}(\theta)$ for some pdf $q_{t}(\theta)$ and agents $i \in \nodes$, 
  the normalization factor $p(z_t|z_{< t})$ in~\eqref{eqn:dBayes} is lower bounded by the separable
  distributed evidence lower bound (DELBO),
  % $\sum_{i \in \nodes} \underset{q_{i,t}(\theta)}{\expect} [\qz{i}(z_{i,t}|\theta) - \frac{1}{n} \log(q_{i,t}(\theta)) 
  % + \sum_{j \in \nodes} \frac{A_{ij}}{n} \log p_{j}(\theta|z_{<t})]$.  
  \begin{equation*}
    \sum_{i \in \nodes} \underset{q_{i,t}}{\expect} [\qz{i}(z_{i,t}|\theta) - \frac{1}{n} \log(q_{i,t}(\theta)) 
  + \sum_{j \in \nodes} \frac{A_{ij}}{n} \log p_{j}(\theta|z_{<t})],
  \end{equation*}  
  where $A$ is the adjacency matrix satisfying Assumption~\ref{assume:connect}.
\end{theorem}

\begin{proof}
Given the agent pdfs $p_{i,t}(\theta| z_{< t})$, the centralized estimate at time $t$ is defined 
as their normalized geometric average $p(\theta| z_{< t}) = \frac{1}{K_{<t}}\prod_{i \in \nodes} 
(p_{i,t}(\theta| z_{< t}))^{1/n}$.
The normalization factor $K_{< t} = \int \prod_{i \in \nodes} 
(p_{i,t}(\theta| z_{< t}))^{1/n} d \theta $ is the integral of the geometric average. 
For the stochastic adjacency matrix in Assumption~\ref{assume:connect}, 
the geometric average satisfies $\prod_{i \in \nodes} 
(p_{i,t}(\theta| z_{< t}))^{1/n} = \prod_{i \in \nodes} (\prod_{j \in \nodes} 
p_{j,t}(\theta| z_{< t})^{A_{ij}} )^{1/n}$. This property relates the agent prior densities with those of the one-hop neighbors. Following the approach for deriving ELBO, the normalization in \eqref{eqn:dBayes} 
is expressed in terms of the agent log likelihoods, priors, and posterior,
\begin{align}
  & \log p(z_t|z_{< t}) = \log \frac{p(z_t|\theta) p(\theta|z_{<t}) }
  {p(\theta| z_{\leq t}) } \\
  & = \log \frac{1}{K_{<t}} \prod_{i \in \nodes} \frac{\qz{i}(z_{i,t}|\theta) p_{i}(\theta|z_{<t})^{1/n} }
  {p(\theta| z_{\leq t})^{1/n} } \\
  & = \log \frac{1}{K_{<t}} \prod_{i \in \nodes} \frac{\qz{i}(z_{i,t}|\theta) \prod_{j \in \nodes} p_{j}(\theta|z_{<t})^{A_{ij}/n} }
  {p(\theta| z_{\leq t})^{1/n} }.
\end{align}
The geometric average of the non-negative pdfs is pointwise upper bounded by their arithmetic average, 
and, hence, its integral satisfies $K_{<t} \leq \int \sum_{j} A_{ij} p_{j,t}(\theta) d\theta = 1$. 
As a result, $\log K_{<t} \leq 0$.
% Change likelihood to \ell
As in the centralized setting, since the argument in pdf $q_{t}(\theta)$ is independent of 
the observation $z_{\leq t}$, the expectation of the normalization factor does not alter its value. 
Assuming that $q_{i,t}(\theta) = q_{t}(\theta)$, we separate the expectation over the agent likelihoods and priors as follows,
\scalebox{0.95}{\parbox{1.\linewidth}{%
\begin{align}
  \label{eqn:c_error}
  & \log p(z_t|z_{< t}) = - \underset{q_{t}(\theta)}{\expect}  \log K_{<t} \\
  & + \underset{q_{t}(\theta)}{\expect}  \sum_{i \in \nodes} 
  \Biggl[ \log \frac{\qz{i}(z_{i,t}|\theta) \underset{j \in \nodes}{\prod} p_{j}(\theta|z_{<t})^{\frac{A_{ij}}{n}} q_{i,t}(\theta)^{\frac{1}{n}}}
  {q_{i,t}(\theta)^{1/n} p(\theta| z_{\leq t})^{1/n} } \Biggr]. \nonumber 
\end{align}
}}
\begin{align}
  \label{eqn:m_error}
  & \log p(z_t|z_{< t}) \geq \sum_{i \in \nodes} \underset{q_{i,t}(\theta)}{\expect} [\log \qz{i}(z_{i,t}|\theta)]  \\
  & + \frac{1}{n} \KL [q_{i,t}(\theta) || p(\theta| z_{\leq t})] - \frac{1}{n} \KL [q_{i,t}(\theta) || p^g_i(\theta|z_{<t})], \nonumber  \\
  & \geq \sum_{i \in \nodes} \underset{q_{i,t}(\theta)}{\expect} [\log \qz{i}(z_{i,t}|\theta)] 
  - \frac{1}{n} \KL [q_{i,t}(\theta) || p^g_i(\theta|z_{<t}) ], \nonumber
\end{align}
where $p^g_i(\theta|z_{<t}) = \prod_{j \in \nodes} p_{j}(\theta|z_{<t})^{A_{ij}}$ in the weighted geometric average of the agent prior pdfs. Since the KL divergence term representing the modeling error 
between the approximation $q_{i,t}$ and the estimate $p(\theta|z_{\leq t})$ is non-negative,
we can drop this term to obtain a separable lower bound of the normalization factor as,
\begin{align}
  & \log p(z_t|z_{< t}) \geq \sum_{i \in \nodes}  \left[ \underset{q_{i,t}(\theta)}{\expect} [\qz{i}(z_{i,t}|\theta)]  \right. \nonumber \\
  & \left. + \frac{1}{n} \sum_{j \in \nodes}\underset{q_{i,t}(\theta)}{\expect} A_{ij} [\log q_{i,t}(\theta) - \log p_{j}(\theta|z_{<t})]
  \right]   \label{eqn:DELBO}
\end{align}
The separable terms contain only the agent's observation $z_i$ 
and are thus analogous to the ELBO at each agent.
\end{proof}

The DELBO derivation in Theorem~\ref{thm:delbo} shows that the posterior approximation error consists of modeling error and consensus error. The consensus error at time $t$ is defined in \eqref{eqn:c_error} as $\log (1/K_{<t})$ 
where $K_{<t} = \int \prod_{i} p_i(\theta|z_{<t})^{1/n} d \theta$. Since $\log (1/K_{<t}) = 1/n \sum_{i\in \nodes} \KL[p_g||p_i(\theta|z_{<t})]$ for $p_g = \prod_{i} p_i(\theta|z_{<t})^{1/n}/K_{<t}$, this error is zero only if the agent pdfs are equal almost everywhere. The modeling error is defined in \eqref{eqn:m_error} as the divergence $\sum_{i} \expect_{q_{i,t}}\KL[q_{i,t} || p_i(\theta|z_{<t})]$. This error is zero only if the pdfs $q_{i,t}$ are computed in the family of accurate posterior densities.
% These errors are induced by our choice of the approximate pdf family $\calF$ and pdf mixing rate from lowest positive weights in $A$ respectively. 
Replacing the accurate pdfs $p_{i}(\theta|z_{<t})$ with their last known approximations~$q_{i,t-1}(\theta)$ 
in family $\calF$ in DELBO yields a separable functional $J_t[q_{1,t},\ldots,q_{n,t}] = \sum_{i \in \nodes} J_{i,t}[q_{i,t}]$ with
\begin{equation*}
  % \label{eqn:Ji}
  J_{i,t}[q_{i,t}] =\!\! \underset{q_{i,t}(\theta)}{\expect} [ \log [\qz{i}(z_{i,t}|\theta) \prod_{j \in \nodes} q_{j,t-1}(\theta)^{\frac{A_{ij}}{n}}] 
  - \log q_{i,t}(\theta)^{\frac{1}{n}}].  
\end{equation*}

\begin{corollary}
  Upon maximizing the DELBO component~$J_{i,t+1}[p]$ of agent $i$, the optimal pdf $q_{i,t+1}(\theta) \in \argmax_p J_{i,t+1}[p]$.
  The optimal pdf satisfies, 
  \begin{equation}
    \label{eqn:geom_update}
    q_{i,t+1} = \qz{i}(z_{i,t+1}|\theta) q^g_i(\theta) \Big/ \int \qz{i}(z_{i,t+1}|\theta) q^g_i(\theta) d\theta, 
  \end{equation}
  where the mixed pdf at agent $i$ is $q^g_i(\theta) = \prod_{j \in \neighbor{i}} q_{j,t}(\theta)^{\frac{A_{ij}}{n}}$
  under the consensus constraint $q_{i,t} = q_{j,t}, \forall i,j \in \nodes$.
\end{corollary}

The weighted sum of KL-divergences in \eqref{eqn:geom_update} penalizes deviation from consensus of the agent pdfs $q_{i,t}(\theta)$. Sharing weighted pdfs with neighbors is key to reaching consistent estimates across the network.
The asymptotic averaging properties $\lim_{t \rightarrow \infty} A^t = \frac{1}{n} \ones{}\ones{}^{\top}$ 
of matrix $A$ generate agent estimates eventually consistent with the centralized one $q_t(\theta) = q_{i, t}(\theta)$.
To observe the impact of matrix $A$ on guaranteeing consensus in distributed estimation problems, please refer to the convergence analysis in \cite{PP-NA-SM:19-cdc, PP-NA-SM:22, AN-AO-CAA:17}.
%In this paper, we will instead focus on variational approximations in distributed setting.

\begin{remark}[Distributed estimation]
  With conjugate agent likelihoods $\qz{i}(z_i|\theta)$ weighted by factor $n$, 
  the distributed updates in \cite{PP-NA-SM:22} match the DELBO updates, 
  thus guaranteeing probabilistic convergence for accurate posterior computations.
\end{remark}

% Rewrite: how does sampling interact with message passing in BayesPy?
The posterior $p(\theta)$ in~\eqref{eqn:geom_update} can be approximated for arbitrary likelihood pdfs 
using black-box VI \cite{RR-SG-DB:14} in the variational message passing framework \cite{JW-CB-TJ:05}.
% We employ this approach in the next example to elucidate the computational complexity of repeated sampling.
We employ this approach in the next example to show the impact of sampling on accuracy.

\begin{figure}[h]
  \centering
  \includegraphics[width=0.5\linewidth]{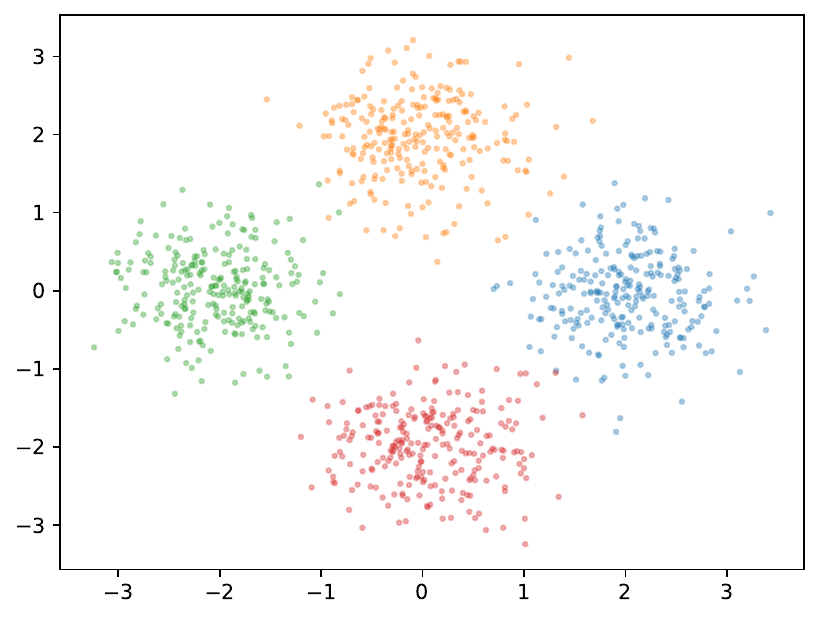}%
  \includegraphics[width=0.5\linewidth]{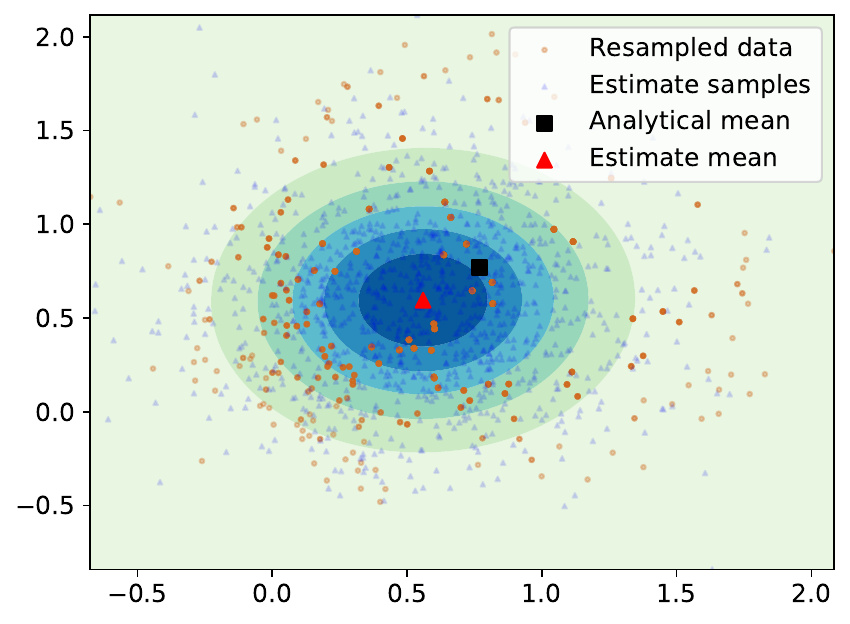}
  \caption{(a) Samples of Gaussian components $p_{i,t}$ centered on a circle of radius $1$ with unit covariance. (b) Particles resampled w.r.t. probability weights computed in \eqref{eqn:geom_update} for data $z_i =[1,1]$, estimated density and analytical mean.}
  \label{fig:glikmix}
\end{figure}

\begin{example}[Estimating geometric mixing of Gaussians]
  \label{ex:mixing}
  In this example, we examine the update in \eqref{eqn:geom_update} for a set of $n=4$ agents with Gaussian priors and likelihoods and observe the update for a single agent that weighs all other agents equally with $A_{ij}=1/n$. 
  Because of sample dependence, we observe that the VI solution to an expressive model may not match the analytical solution. 
  Assume that the Gaussian priors are $p_{i,t}(\theta) = \mathcal{N}(\mu_{i,t}, (\Omega_{i,t})^{-1})$ with means $\mu_{i,t}$, and information matrices $\Omega_{i,t}$. Suppose that the local observation likelihoods $\ell_i(z_{i,t} | \theta) = \mathcal{N}(H \theta, (\Omega_{i}^z)^{-1})$ are Gaussian as well. 
  Since the geometric average of the priors is conditionally conjugate to the likelihood, the posterior at agent~$i$ is
  $\mathcal{N}( \Omega_{i,t+1}^{-1}(H^{\top}\Omega_{i}^z z_i + \sum_{j=1}^n A_{ij} \Omega_{j,t} \mu_{j,t}), \Omega_{i,t+1}^{-1})$, 
  with information matrix $\Omega_{i,t+1} = H^{\top}\Omega_{i}^z H + \sum_{j=1}^n A_{ij} \Omega_{j,t}$. 
  Next, we estimate this Gaussian posterior using VI with sampling \cite{JL:21}. 
  Let the agent estimate an expressive pdf $p(\theta) = \calN(\theta | \mu, \Omega^{-1})\, p_\mu \, p_{\Omega}$ using observation $z_i$ 
  and prior normal distribution $p_\mu = \calN(\mu_p, \Sigma_p)$ on the mean $\mu$ and Wishart distribution $p_{\Omega} = W(\lambda, V)$ 
  on the precision matrix. 
  % The normal distribution has mean $\mu_p$ and covariance $\Sigma_p$, 
  % and the Wishart distribution has degree of freedom $\lambda$ and scale matrix $V$.
  To estimate $p_{i,t+1}(\theta)$ with $p(\theta)$, we consider the component pdfs 
  $p_{i,t}$ as the proposal for generating samples on~$\theta$ and weigh each sample 
  with $\qz{i}(z_{i}|\theta)\prod_{j \in \neighbor{i}} q_{j,t}(\theta)^{\frac{1}{n}}$ 
  from the update in \eqref{eqn:geom_update}. 
  Upon normalization, stratified resampling generates samples representing the posterior
  which is then used to obtain $p(\theta)$.  
  With significant sampling, the mean and covariance of the density inferred in Fig.~\ref{fig:glikmix}
  is similar to the resampled particles.
  Since the VI objective in this example depends on the sampled particles, a minor discrepancy is observed in the estimated mean and analytical value. Although this works fine for a single estimate, it becomes computationally expensive in high-frequency online estimation settings such as filtering.
  Therefore, we will develop approximate analytical updates to perform online inference. 
\end{example}

% Numbers and legend sizes
 
In this section, we derived a distributed variational inference algorithm in~\eqref{eqn:geom_update}
requiring costly computation of the normalization factor.
To enable efficient implementation, we further develop this algorithm to use stochastic gradients
of log-likelihood terms and compute their analytical approximations. 

% \begin{example}[Comparison with extended Kalman filter]
%   Figure~\ref{fig:estimates} presents the results for a source localization problem, where the sensors receive 
%   noisy distances to a source with unknown location $\theta^\star$. 
%   The range likelihood model for agent $i$ is $\qz{i}(z_i|\theta) = \phi(z_i|\Vert x_i - \theta \Vert_2, 1)$
%   in terms of agent positions $x_i$.
%   At each time step, the agents update their posterior estimates
%   using distributed extended Kalman filter % A reference maybe
%    and sample based VI. 
% \begin{figure}[h]
%   \label{fig:estimates}
%   \centering
%   \includegraphics[width=0.24\textwidth]{figures/source_1_A.png}
%   \includegraphics[width=0.24\textwidth]{figures/source_1_VI.png}
%   % \includegraphics[width=0.32\textwidth]{figures/source_particle_estimation.png}
%   \caption{(a) Distributed extended Kalman filter, and (b) distributed variational inference 
%   estimation in source localization problem with true location $\theta^\star = [1,2]$.}
%   % , and (c) distributed variational inference sampling with posterior based proposal sampling.
% \end{figure}
% \end{example}

%% file: section5_dgvi.tex
%%%%%%%%%%%%%%%%%%%%%%%%%%%%%%%%%%%%%%%%%%%%%%%%%%%%%%%%%%%%%%%%%%%%%%%%%%
\section{Distributed Gaussian variational inference}
\label{sec:dgvi}
%%%%%%%%%%%%%%%%%%%%%%%%%%%%%%%%%%%%%%%%%%%%%%%%%%%%%%%%%%%%%%%%%%%%%%%%%%
This section derives agent specific iterative updates for variational inference with Gaussian variational densities and arbitrary log-likelihood functions. Appropriate approximations to the expected log-likelihood derivatives are devised to generate analytical Gaussian updates for distributed classification and regression problems. Further, rank-correcting inverse and diagonalized covariance updates are presented to support real-time implementation.

%%%%%%%%%%%%%%%%%%%%%%%%%%%%%%%%%%%%%%%%%%%%%%%%%%%%%%%%%%%%%%%%%%%%%%%%%%
\subsection{Distributed Gaussian variational inference (DGVI)}
%%%%%%%%%%%%%%%%%%%%%%%%%%%%%%%%%%%%%%%%%%%%%%%%%%%%%%%%%%%%%%%%%%%%%%%%%%

We assume that the agents collect observations from arbitrary likelihoods but restrict their variational pdfs $q_{i,t}(\theta)$ to a Gaussian pdf family $\mathcal{F}$. The solution to the ELBO optimization in \eqref{eqn:c_geom_update} for a Gaussian pdf family $\mathcal{F}$ is stated in the next lemma.

%We assume that the agents collect data based on arbitrary
%likelihoods, but estimate a density on the parameter $\theta$ in the family of Gaussian pdfs.
%As shown in~\eqref{eqn:ELBO}, the ELBO term for centralized case is 
%$\expect_{q_t(\theta)} [\qz{}(z_t|\theta) + \log q_{t-1}(\theta) - \log q_{t}(\theta)]$.
%The solution to this ELBO for a Gaussian $q_t$ yields Lemma~\ref{lemma:gvis}.

\begin{lemma}[Gaussian variational inference]\label{lemma:gvis} 
  Assume that the known prior density $q_{t-1}(\theta)$ is a Gaussian $\calN(\theta|\mu_{t-1}, \Omega_{t-1}^{-1})$ with mean $\mu_{t-1}$ and information matrix $\Omega_{t-1}$. Then, the Gaussian pdf $q_{t}$ minimizing the ELBO in \eqref{eqn:c_geom_update} is,
\begin{align}
  \label{eqn:gvi_sample}
  \begin{split}
    \Omega_{t} & = \Omega_{t-1} -  \expect_{q_{t-1}} [\nabla_{\theta}^2 \log \qz{} (z_{t}|\theta)], \\
    \mu_{t} & = \mu_{t-1} + \Omega_{t}^{-1} \expect_{q_{t-1}} [\nabla_{\theta} \log \qz{} (z_{t}|\theta)].
  \end{split}
\end{align}
\end{lemma}

\begin{proof}
  The proof is presented in Appendix~\ref{app:secgaussianalgo}. We pose the ELBO objective as the loss functional in \cite[Eqn.~25]{TDB-JRF-DJY:20}, which avoids the implicit expectation of the form $\expect_{q_{t}} [\nabla_{\theta} \log \qz{} (z_{t}|\theta)]$ as seen in \cite{ML-SB-FB:22}.
% Their proof applies Taylor expansion and Stein's lemma to this .
%It avoids the implicit expectation terms for the derivation in \cite{ML-SB-FB:22}.
% A similar derivation in \cite{ML-SB-FB:22} is avoided as the update optimizes an implicit expectation term.
\end{proof}

% A further simplification yields,
% \begin{align}
%   \mu_t & = \mu_{t-1} + \Omega_{t-1}^{-1} \nabla_{\mu_t} \expect_{q_t} [\log p (z_t|\theta)] \\
%   \Omega_t & = \Omega_{t-1} - 2 \nabla_{\Omega_t^{-1}} \expect_{q_t} [\log p (z_t|\theta)]
% \end{align} 

%\paragraph*{Distributed version}
The DELBO in Theorem~\ref{thm:delbo} admits separable objectives for each agent, 
such that each DELBO component contains only the agent's observation model and neighbor priors.
Lemma~\ref{lemma:gvis} has an online update minimizing the ELBO objective 
over the set of Gaussian densities in $\calF$.
The following lemma solves the agent-component of the distributed optimization problem 
in \eqref{eqn:geom_update} over Gaussian densities.

\begin{lemma}[Distributed Gaussian variational inference]\label{lemma:dgvis}
Assume that agent $i$ receives observation $z_{i,t+1}$ with likelihood $\qz{}(z_{i,t+1}|\theta)$ and neighbor estimates $q_{j,t}(\theta) = \calN(\theta|\mu_{j,t}, \Omega_{j,t}^{-1})$ at time $t$. 
Upon weighing neighbor opinions with elements of matrix $A$,
the mean $\mu_{i,t+1}$ and information matrix $\Omega_{i,t+1}$ of the pdf $q_{i,t+1}$ minimizing DELBO in \eqref{eqn:geom_update} is,
  \begin{align}
    \label{eqn:dgvi_sample}
    &\Omega_{i,t+1}^g = \sum_{j \in \nodes} A_{ij} \Omega_{j, t}, \Omega_{i,t+1}^g \mu^g_{i,t+1} = %()^{-1} 
    \sum_{j \in \nodes} A_{ij} \Omega_{j, t}\mu_{j, t}\nonumber \\ 
    &\Omega_{i,t+1} = \Omega_{i,t+1}^g -  \expect_{q_{t}^g} [\nabla_{\theta}^2 \log \qz{} (z_{i,t+1}|\theta)],  \\
    & \mu_{i,t+1} = \mu_{i,t+1}^g + (\Omega_{i,t+1}^g)^{-1} \expect_{q_{i,t}^g} [\nabla_{\theta} \log \qz{} (z_{i,t+1}|\theta)]. \nonumber 
  \end{align}
\end{lemma}

\begin{proof}
The mean $\mu^g_{i,t+1}$ and information matrix $\Omega^g_{i,t+1}$ 
of the weighted geometric average of Gaussians is given in \cite{PP-NA-SM:22}.
The remainder follows from the proof of Lemma~\ref{lemma:gvis}. 
\end{proof}

Both the centralized and distributed Gaussian variational update rules in Lemmas~\ref{lemma:gvis} and~\ref{lemma:dgvis} require the expected  log-likelihood gradient and Hessian terms. Their estimation using Monte Carlo methods is computationally expensive, especially for high-dimensional parameters. We obtain analytic approximations of the gradient and Hessian expectations for classification and regression problems in the next two subsections.

%Both these lemmas contain expected likelihood gradient and Hessian terms. 
%Their estimation using Monte Carlo methods is computationally expensive for high-dimensional parameters. 
%Therefore, we next obtain analytic approximations of the expectation for classification and regression problems.

%%%%%%%%%%%%%%%%%%%%%%%%%%%%%%%%%%%%%%%%%%%%%%%%%%%%%%%%%%%%%%%%%%%%%%%%%%
\subsection{DGVI for classification}
\label{sec:dgvi_classification}
%%%%%%%%%%%%%%%%%%%%%%%%%%%%%%%%%%%%%%%%%%%%%%%%%%%%%%%%%%%%%%%%%%%%%%%%%%
We consider a kernel-based observation likelihood model for probabilistic classification.
The kernel parameters consist of a set of known fixed feature points and corresponding weights.
%The free weight parameters are optimized as per the received data.
The data $z = (x, y)$ is embedded in feature space by a transformation $\Phi_x = [1, k_1(x), \dots, k_l(x)]$ with
elements $k_s(x) = \gamma_1 \exp(-\gamma_2 \Vert x - x^{(s)} \Vert^2)$ where $x^{(s)}$ are the known kernel centers and $(\gamma_1, \gamma_2)$ are kernel scaling parameters chosen to suit the domain and regularity of the model. The likelihood of an observation $z = (x,y)$ with input $x \in \real^d$, feature $\Phi_x \in \real^{l+1}$, and label $y \in \{0,1\}$ is modeled as,
\begin{equation}\label{eqn:observation_model_sig}
  \qz{}(z | \theta) = \sigma(\Phi_x^{\top} \theta)^y (1-\sigma(\Phi_x^{\top} \theta))^{1-y},
\end{equation}
where $\theta$ are the model parameters and $\sigma$ is the sigmoid function.

%to model output $y = \sigma(\Phi_x^{\top} \theta) = 1/(1+\exp(-\Phi_x^{\top} \theta))$. 
%The kernel centers $x^{(i)}$ and the scaling parameters $(\gamma_1, \gamma_2)$ are
%chosen to suit the domain and regularity of the model.
%For the dataset,
%\begin{align*}
%  \calD = \{z_m\}_{m=1}^M, z_m = (x, y), x \in \real^d, y \in \{0,1\},
%\end{align*}
%the class observation likelihood is modeled as,
%\begin{align}
%  \label{eqn:observation_model_sig}
%  \qz{}(z_m| \theta) = \sigma(\Phi_x^{\top} \theta)^y (1-\sigma(\Phi_x^{\top} \theta))^{1-y}.
%\end{align}

% We will next attempt to compute analytical approximations of these terms.

To estimate the distribution of the parameters $\theta$ using the GVI algorithm in Lemma~\ref{lemma:gvis}, 
we would need to estimate the expectation over the log-likelihood gradient, $\nabla_\theta \log p(z| \theta)$, and Hessian, $\nabla_\theta^2 \log p(z| \theta)$. We derive an analytical approximation to these terms. With $\nabla_\theta \sigma(\Phi_x^{\top}\theta) = \sigma(\Phi_x^{\top}\theta) (1- \sigma(\Phi_x^{\top}\theta)) \Phi_x^{\top}$,
the log-likelihood derivatives are, 
\begin{align}
  & \log \qz{}(z| \theta) = y \log \sigma(\Phi_x^{\top} \theta) + (1-y) \log (1-\sigma(\Phi_x^{\top} \theta)), \nonumber\\
  % = \left[ y \sigma(\Phi_x^{\top} \theta) \exp(-\Phi_x^{\top} \theta) 
  % - (1-y)\frac{\sigma(\Phi_x^{\top} \theta)^2}{1-\sigma(\Phi_x^{\top} \theta)}  \exp(-\Phi_x^{\top} \theta)\right] \Phi_x^{\top} \\
  \label{eqn:loglikgrad_sigmoid}
  & \nabla_\theta \log \qz{}(z| \theta) 
  = (y - \sigma(\Phi_x^{\top} \theta)) \Phi_x^{\top}, \\
  % & = \frac{\sigma(\Phi_x^{\top} \theta) (y - \sigma(\Phi_x^{\top} \theta))}{1-\sigma(\Phi_x^{\top} \theta)} \exp(-\Phi_x^{\top} \theta) \Phi_x^{\top} \\
  & \nabla_\theta^2 \log \qz{}(z| \theta)
   = - \sigma(\Phi_x^{\top}\theta) (1- \sigma(\Phi_x^{\top}\theta)) \Phi_x\Phi_x^{\top} .
  % \\
  % & \frac{\sigma(\Phi_x^{\top} \theta)^2 (y - \sigma(\Phi_x^{\top} \theta))}{1-\sigma(\Phi_x^{\top} \theta)} \exp(-2 \Phi_x^{\top} \theta) \Phi_x \Phi_x^{\top}
  % - \frac{\sigma(\Phi_x^{\top} \theta)^3 }{1-\sigma(\Phi_x^{\top} \theta)} \exp(-2\Phi_x^{\top} \theta) \Phi_x \Phi_x^{\top} \\
  % & - \frac{\sigma(\Phi_x^{\top} \theta) (y - \sigma(\Phi_x^{\top} \theta))}{1-\sigma(\Phi_x^{\top} \theta)} \exp(-\Phi_x^{\top} \theta) \Phi_x\Phi_x^{\top} 
  % + \frac{\sigma(\Phi_x^{\top} \theta)^3 (y - \sigma(\Phi_x^{\top} \theta))}{2(1-\sigma(\Phi_x^{\top} \theta))^2} \exp(-\Phi_x^{\top} \theta) \Phi_x\Phi_x^{\top} 
\end{align}
% http://eelxpeng.github.io/blog/2017/03/10/Tricks-of-Sigmoid-Function
To analytically compute the expectation of gradient, Hessian and their derivative terms with respect to a Gaussian density, 
we approximate the sigmoid function $\sigma(x)$ with an inverse probit function 
$\Gamma(\xi x) = \int_{-\infty}^{\xi x} \phi(\alpha|0,1) d\alpha$ for $\xi = 0.61$ according to \cite{JD:17}.
Fortunately, the expectation of the inverse probit function with respect to a Gaussian density is an inverse probit. For the second derivative, the derivative of the sigmoid function is approximated via a Gaussian probability density function $\phi$ with zero mean and unit covariance. Using $\sigma(\Phi_x^{\top}\theta) \approx \Gamma(\xi \Phi_x^{\top}\theta)$, the Hessian becomes,  
\begin{align}
  \nabla_\theta^2 \log \qz{}(z| \theta) & = - \nabla_\theta \sigma(\Phi_x^{\top} \theta) \Phi_x^{\top}  
    \approx - \nabla_\theta \Gamma(\xi \Phi_x^{\top} \theta) \Phi_x^{\top} \nonumber \\
    \label{eqn:loglikhess_sigmoid}
    & = - \xi \phi (\xi \Phi_x^{\top} \theta| 0, 1) \Phi_x \Phi_x^{\top}.
\end{align}

The DGVI algorithm in Lemma~\ref{lemma:dgvis} contains the expectation over gradient and Hessian terms,
that we approximate next.
\begin{lemma}[Expected log-likelihood gradient and Hessian]\label{lemma:gradient}
  For probabilisitic classification with a kernel-based observation likelihood model in \eqref{eqn:observation_model_sig},
  the expected gradient and Hessian of the log-likelihood in \eqref{eqn:loglikgrad_sigmoid} 
  with respect to a Gaussian density $q_t(\theta) = \phi(\theta| \mu_t, \Omega_t^{-1})$ satisfy,
  \begin{align}
    & \expect_{q_{t}}[\nabla_\theta \log \qz{}(z| \theta) ] \approx \left( y - \Gamma 
    \left( \frac{\xi \Phi_x^{\top} \mu_{t}}{\sqrt{\beta}} \right) \right) \Phi_x^\top, \nonumber \\
    & \underset{q_{t}}{\expect} [\nabla_{\theta}^2 \log \qz{} (z_{t+1}|\theta)] \\
    & \approx -  \sqrt{\frac{\xi^2}{2 \pi \beta}}
    \exp\left(-\frac{1}{2} [\frac{\xi^2}{\beta} \mu_t^{\top} \Phi_x \Phi_x^{\top} \mu_t] \right)\Phi_x\Phi_x^{\top} , \nonumber
  \end{align}
  where $\beta = 1+\xi^2 \Phi_x^{\top} \Omega_{t}^{-1} \Phi_x$.
\end{lemma}

\begin{proof}
Please refer to Appendix~\ref{app:expect}.
\end{proof}

Methods to estimate Gaussian variational posteriors are surveyed in \cite{HN-CER:08}, 
and the expectation propagation method is recommended for its accuracy. However, the associated 
computational complexity may not allow real-time implementation. Our approximations of the log-likelihood gradient and Hessian expectations can be substituted in Lemma~\ref{lemma:dgvis} to obtain analytical updates for approximate distributed Gaussian VI. In the distributed setting, each agent knows the fixed kernel centers $\{x^{(s)}\}$ and scale parameters $\gamma_1, \gamma_2$, receives private observations $z_{i,t}$, and estimates a pdf over the weights $\theta$.

\begin{lemma}[DGVI for kernel classification]\label{lemma:dgvi_gaussian}
For observation $z=(x, y)$ received at agent $i$, classification likelihood defined in \eqref{eqn:observation_model_sig}, and neighbor estimates $\phi(\theta| \mu_{j,t}, \Omega_{j,t}^{-1})$, the DELBO maximizing Gaussian density $q_{i,t}(\theta) = \phi(\theta| \mu_{i,t}, \Omega_{i,t}^{-1})$ is,
  \begin{align}
    \label{eqn:dgvi_classify}
    \Omega_{i,t}^g & = \sum_{j \in \nodes} A_{ij} \Omega_{j, t}, \, \Omega_{i,t}^g \mu^g_{i, t} = %()^{-1} 
    \sum_{j \in \nodes} A_{ij} \Omega_{j, t}\mu_{j, t}, \nonumber \\
    \Omega_{i,t+1} & = \Omega_{i,t}^g +  \gamma \Phi_x \Phi_x^{\top}, \\
    \label{eqn:domega_inv}
    \Omega_{i,t+1}^{-1} & = (\Omega_{i,t}^g)^{-1} - \gamma/\gamma_1 
    (\Omega_{i,t}^g)^{-1} \Phi_x\Phi_x^{\top} (\Omega_{i,t}^g)^{-1} \\
    \mu_{i,t+1} & = \mu_{i,t}^g + \left(y - \Gamma \left( 
      \frac{\xi \Phi_x^{\top} \mu_{i,t}^g}{\sqrt{\beta}}\right)\right) \Omega_{i,t+1}^{-1} \Phi_x 
  \end{align}
  with $\beta = 1 + \xi^2 \Phi_x^{\top} (\Omega_{i,t}^g)^{-1} \Phi_x$, $\gamma_1 = 1 + \gamma\Phi_x^{\top} (\Omega_{i,t}^g)^{-1} \Phi_x$
  and $\gamma = \sqrt{\frac{\xi^2}{2 \pi \beta}} 
  \exp\left(-0.5 [\frac{\xi^2}{\beta} (\mu_{i,t}^g)^{\top} \Phi_x \Phi_x^{\top} \mu_{i,t}^g] \right)$.
\end{lemma}

\begin{proof}
  The mean $\mu^g_{i,t}$ and information matrix $\Omega^g_{i,t}$ 
  represents the geometric average of prior Gaussians.
  For the rest, we compute the Gaussian minimizing the agent separable bound DELBO using the steps for Lemma~\ref{lemma:dgvis}. 
  The expected gradients are derived in the proof for Lemma~\ref{lemma:gradient} followed by steps reducing matrix inversion computations
  in Appendix \ref{app:expect}. 
  % It uses probit approximation of the expected gradient and Hessian terms
  % followed by matrix determinant lemma and Woodbury's lemma for further simplification\NA{It is not necessary to resate the proof of Lemma~\ref{lemma:gradient} here.}.  
\end{proof}

The DGVI updates in Lemma~\ref{lemma:dgvi_gaussian} include two linear system solutions $(\Omega_{i,t}^g)^{-1} (\sum_{j \in \nodes} A_{ij} \Omega_{j, t}\mu_{j, t})$ and $(\Omega_{i,t}^g)^{-1} \Phi_x$.  
In a centralized setting, the matrix inversion needs to be performed only at 
the first step to compute $\Omega_0^{-1}$, and any following inverses may be computed iteratively in \eqref{eqn:domega_inv}.
The costly matrix inversion can be avoided by using Gaussian variational densities 
with diagonal covariances, which we discuss next.
% Since the inversion of arithemetic average of the information matrix is computationally costly,
% we can consider the geometric average of these matrices as discussed in \cite{XY-WH-PA-KG:20}.
% \begin{align*}
%   \bar{\Omega}_{ii,t} = A_{ii} \Omega_{i,t}, \bar{\Omega}_{ij,t} = A_{ij} \Omega_{j,t},
%   \Omega_{i,t+1}^g = \bar{\Omega}_{ii,t}(I + (1/A_{ii})\Omega_{i,t}^{-1}\sum_{j \in \neighbor{i} \backslash i} 
%   \bar{\Omega}_{ij,t})
% \end{align*}

% For the observation model defined in the classification problem in
% measurement $(x, y)$ received at agent $i$, diagonalized covariance 
  
\begin{lemma}[Diagonalized GVI for kernel classification] \label{lemma:diag_dgvi}
  For observation $z = (x, y)$ received at agent $i$, classification likelihood defined in \eqref{eqn:observation_model_sig}, and neighbor estimates $\phi(\theta| \mu_{j,t}, D_{j,t}^{-1})$ with diagonal information matrices $D_{j,t}$, the iterative GVI update to Gaussian density $q_t(\theta) = \phi(\theta| \mu_{i,t}, D_{i,t}^{-1})$ with diagonal information matrix $D_{i,t}$ is,
  \begin{align}
    & D_{i,t}^g = \sum_{j \in \nodes} A_{ij} D_{j, t}, \;\; \mu^g_{i, t} = (D_{i,t}^g)^{-1} 
    \sum_{j \in \nodes} A_{ij} D_{j, t}\mu_{j, t}, \nonumber \\
      & D_{i,t+1}  %= - \expect_{q_{t}} [\nabla_{\theta}^2 \log p (z_{t+1}|\theta)] \\
      = D_{i,t}^g + \gamma \sqrt{\xi^2 / 2 \pi \beta}   \, 
      \mathrm{diag}(\Phi_x \Phi_x^{\top}), \\
      & \mu_{i,t+1} = \mu_{i,t}^g + (D_{i,t}^g)^{-1} \left( y - \Gamma 
      \left(\frac{\xi \Phi_x^{\top} \mu_{i,t}^g}{\sqrt{\beta}} \right) \right) \Phi_x^\top, \nonumber
  \end{align}
  where $\gamma = \exp\left(-0.5 [\frac{\xi^2}{\beta} (\mu_{i,t}^g)^{\top} \Phi_x \Phi_x^{\top} \mu_{i,t}^g] \right)$, and 
  $\beta = 1+\xi^2 \Phi_x^{\top} (D_{i,t}^g)^{-1} \Phi_x$.
\end{lemma}

\begin{proof}
  The mean $\mu^g_{i,t}$ and information matrix $D^g_{i,t}$ of the 
  geometric average of Gaussians is given in Lemma~\ref{lemma:dgvis}. 
  Please refer to Appendix \ref{app:diag_G} for the remainder.
  %Since the information matrix is diagonal, covariance computation is cheap.
\end{proof}

%%%%%%%%%%%%%%%%%%%%%%%%%%%%%%%%%%%%%%%%%%%%%%%%%%%%%%%%%%%%%%%%%%%%%%%%%%%%%%%%%%%%%%%%%%
\subsection{Distributed Gaussian variational inference for regression}
%%%%%%%%%%%%%%%%%%%%%%%%%%%%%%%%%%%%%%%%%%%%%%%%%%%%%%%%%%%%%%%%%%%%%%%%%%%%%%%%%%%%%%%%%%
In this section, we derive distributed Gaussian VI updates for regression. Consider a linear model $y = \Phi_x^{\top} \theta$ defined using a feature vector $\Phi_x = [1, k_1(x), \dots, k_l(x)]$ with elements $k_m(x)$ defined as in Sec.~\ref{sec:dgvi_classification} and parameters $\theta$. Assume that agent $i$ receives observation $z_i=(x, y)$ sampled from $\qz{i}(z_i|\theta) \propto \exp(-0.5(y - \Phi_x^{\top} \theta)^{\top}S_i(y - \Phi_x^{\top} \theta))$ with symmetric and positive definite $S_i = S_i^{\top}$.

%Now, we can discuss the other aspect of supervised learning, that is regression for learning outputs in continuous spaces.
%Let us consider a linear model $y = \Phi_x^{\top} \theta$ defined using a set of arbitrary 
%kernel functions $\Phi_x = [1, k_1(x), \dots, k_l(x)]$, 
%that best represent the relationship between input and decision variables. 
%We assume that agent $i$ in the network $\graph$ receives observation $z_i=(x, y)$ 
%sampled from $\qz{i}(z_i|\theta) = \exp(-0.5(y - \Phi_x^{\top} \theta)^{\top}S_i(y - \Phi_x^{\top} \theta))$ 
%with symmetric $S_i = S_i^{\top}$.
%In a distributed setting, each agent receives data $z_i$ privately to locally estimate a  
%pdf $q_i(\theta)$ over the unknowns $\theta$.

\begin{lemma}[DGVI for kernel regression]
  \label{lemma:gvi_mvr}  
  Assume that agent $i$ receives data $(x,y)$ and neighbor estimates $\phi(\theta| \mu_{j,t}, \Omega_{j,t}^{-1})$ to learn the Gaussian density 
  $q_{i,t+1}(\theta) = \phi(\theta| \mu_{i,t+1}, \Omega_{i,t+1}^{-1})$.
  The Gaussian $q_{i,t+1}(\theta)$ maximizing DELBO for regression is, 
  \begin{align}
    & \Omega_{i,t}^g = \sum_{j \in \nodes} A_{ij} \Omega_{j, t}, 
     \mu^g_{i, t} = (\Omega_{i,t}^g)^{-1} \sum_{j \in \nodes} A_{ij} \Omega_{j, t}\mu_{j, t}\\
      & \Omega_{i,t+1} = \Omega_{i,t}^g + \Phi_x S_i \Phi_x^{\top}, \Sigma_{i,t}^g = (\Omega_{i,t}^g)^{-1}\\
      & \Omega_{i,t+1}^{-1} = \Sigma_{i,t}^g - \Sigma_{i,t}^g \Phi_x 
      (S_i^{-1} + \Phi_x^{\top} \Sigma_{i,t}^g \Phi_x)^{-1} \Phi_x^{\top} \Sigma_{i,t}^g \nonumber \\
      & \mu_{i,t+1} = \mu_{i,t}^g + (\Omega_{i,t+1})^{-1} (\Phi_x S_i^{\top} y - \Phi_x S_i \Phi_x^{\top} \mu_{i,t}^g) 
  \end{align}
\end{lemma}

\begin{proof}
  Please refer to Appendix~\ref{app:regression}.
\end{proof}
% This update avoids inverting the model dimensional matrix, 
%   instead choosing to invert much smaller data dimensions ($d_y << l$).
%%%%%%%%%%%%%%%%%%%%%%%%%%%%%%%%%%%%%%%%%%%%%%%%%%%%%%%%%%%%%%%%%%%%%%%%%%%%%%%%%%%%%%%%%%

%% file: section6_results.tex
%%%%%%%%%%%%%%%%%%%%%%%%%%%%%%%%%%%%%%%%%%%%%%%%%%%%%%%%%%%%%%%%%%%%%%%%%%%%%%%%%%%%%%%%%%
\section{Results}
\label{sec:results}
%%%%%%%%%%%%%%%%%%%%%%%%%%%%%%%%%%%%%%%%%%%%%%%%%%%%%%%%%%%%%%%%%%%%%%%%%%%%%%%%%%%%%%%%%%
In this section, we evaluate our distributed inference algorithms on classification and mapping datasets. 
For mapping, the functions $\Phi_x$ in \eqref{eqn:observation_model_sig} are kernel functions rooted around the spatial point $x^{(i)}$, and corresponding $\theta_i$ represent the weight on the corresponding occupancy kernel.
We first use this model to perform centralized inference for binary classification on a toy dataset. Then, we demonstrate distributed inference for probabilistic occupancy mapping using two LiDAR datasets.\footnote{Source code available at \url{https://github.com/pptx/distributed-mapping}.}

%We compare the effect of maintaining full covariance, generate probablilistic maps, and observe the learned parameters in relevance vector machines. 

\paragraph*{Toy data}

\begin{figure}[h]
  \centering
  \includegraphics[width=0.51\linewidth]{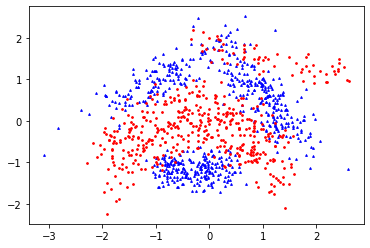}%
  \includegraphics[width=0.485\linewidth]{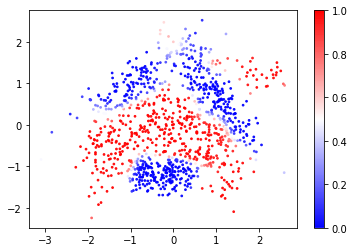}
  \caption{True point classes for Banana dataset (left) and predicted probability $\expect_{q(\theta_T)}p(x,y|\theta)$ of point $(x, y)$ belonging to the red class (right).}
  \label{fig:reg_estimates}
\end{figure}

We consider the Banana dataset \cite{AB-SE-JW:05}, which consists of $5300$ points with binary labels, visualized in Fig.~\ref{fig:reg_estimates}. The probability of each point belonging to the first class, estimated by centralized version of our VI algorithm in Lemma~\ref{lemma:dgvi_gaussian}, is also visualized in Fig.~\ref{fig:reg_estimates}. We pick $50$ feature points at random, with scale $\gamma_1 = 1$ and lengthscale $\gamma_2 = 0.3$
to construct feature functions $\Phi_x$ as defined prior to~\eqref{eqn:observation_model_sig}. We select $50\%$ data for training, and run the single-agent version of the algorithm in Lemma~\ref{lemma:dgvis} updating the mean and covariance of the weights $\theta$ over the feature points.
With $20k$ steps, the algorithm achieves $88\%$ classification accuracy on test set.

%We start by first presenting the results on a small artificial dataset. 
%The Banana dataset, also presented in \cite{AB-SE-JW:05}, consists of just $5300$ instance of binary class data. 
%The interspersed clusters from each class emphasize the need for probabilistic prediction. 
%In Fig.~\ref{fig:reg_estimates}, 
%the points are presented as per their class. In the right figure, the expected probability of each point
%belonging to either set is follows from the associated colormap. 
%We pick $50$ feature points at random, with a lengthscale $\gamma_2 = 0.3$, 
%$50\%$ data for training, and run the algorithm for $20k$ steps to achieve $88\%$ prediction accuracy.

% Explain distributed mapping, the datasets, the RVM choice and the individual figures.
\paragraph*{Intel LiDAR dataset \cite{AH-NR:03}} In a cooperative mapping problem, robots 
follow their own trajectories and cooperate to infer a common map of the environment. A LiDAR sensor uses time of flight information to compute the distance to obstacles in several directions. To construct a dataset from this distance information, the points along the rays connecting the robot to obstacles are sorted into free and occupied points \cite{TD-MY-NA:22}. We assume that each robot in the network collects occupancy information in the form of this binary data from the LiDAR scans along its trajectory. To reduce the mapping effort, the robot trajectories may cover disjoint portions of the observed space, generating local data with different distributions.
%, generating non-iid data.

\begin{figure}[h]
  \centering
  \includegraphics[width=0.49\linewidth]{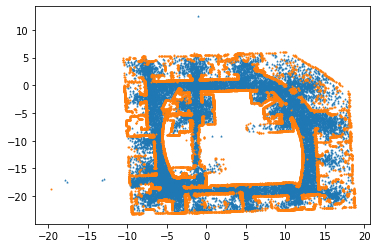}
  \includegraphics[width=0.49\linewidth]{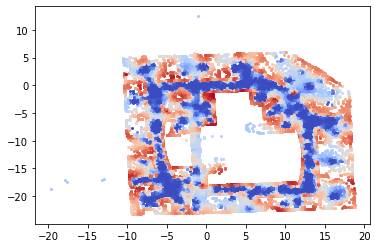}
  \caption{True point classes are presented with orange for occupied spaces as collected by LiDAR scans (left). 
  % Predicted point classes indicating free spaces in blue color ($\expect_{q(\theta_T)}p(x,y|\theta) < 0.5$) using a model with $1200$ feature points.
  Predicted occupancy probability $\expect_{q(\theta_T)}p(x,y|\theta)$ at position $(x,y)$ in the test set.
  The darker red colors represent high occupancy probability, whereas 
blue represents the free space.}
  \label{fig:intel_true_pred}
\end{figure}

% where we use the complete data set
Fig.~\ref{fig:intel_true_pred} presents the results for single agent version of the algorithm in Lemma~\ref{lemma:dgvi_gaussian}. We use $90\%$ of the dataset for training. The remainder forms the test set with a small subset of $1000$ samples forming the verification set for calculating the runtime error. The model is generated using $1200$ feature points selected randomly from the testing set, with scale $\gamma_1 =1$ and lengthscale $\gamma_2 = 0.5$. The diagonalized version of the algorithm in Lemma~\ref{lemma:diag_dgvi} runs for $400k$ steps to achieve $87\%$ accuracy on the test set.

\begin{figure}[h]
  \centering
  \includegraphics[width=0.48\textwidth]{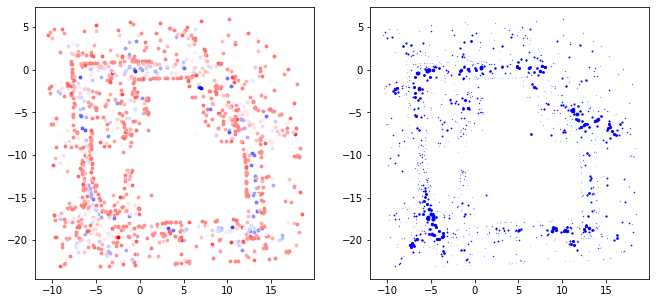}
  \caption{
  Mean $\mu_T$ and variance $\Sigma_T$ of the parameter $\theta$ on $1200$ feature points. 
  Owing to the relevance vector model definition, the mean 
  and variance represent the effect of the estimate at the spatial point on the final prediction.}
  \label{fig:slam}
\end{figure}

Fig.~\ref{fig:slam} presents details on model parameters and probabilistic outputs on the test set.
The lower two images present the mean and diagonal covariance value at the individual feature points 
selected in the map. The right image presents the variance associated with the estimated weight at each of the features. Higher variance is observed at the boundary of the 
free and occupied spaces.

\begin{figure}[h]
  \centering
  \includegraphics[width=0.5\linewidth]{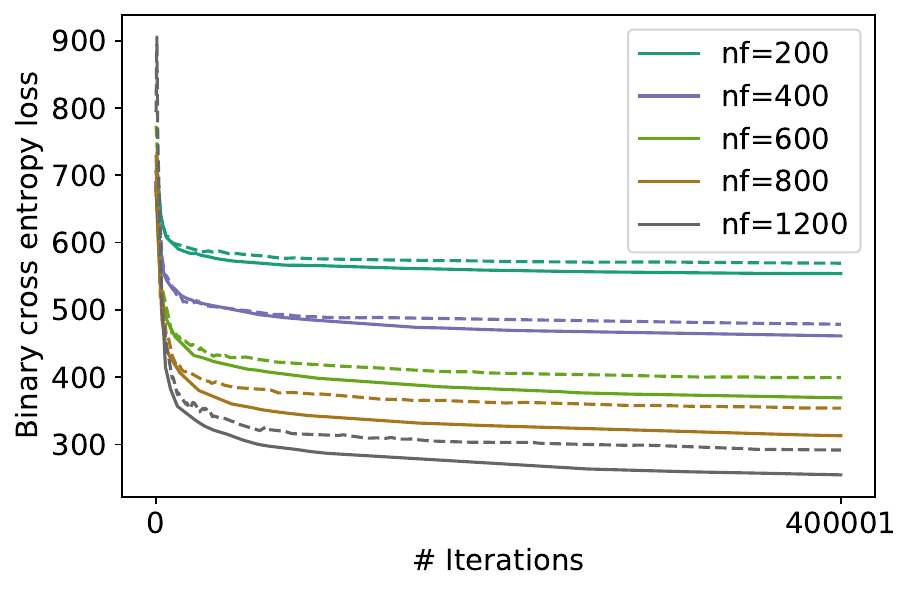}
  \includegraphics[width=0.485\linewidth]{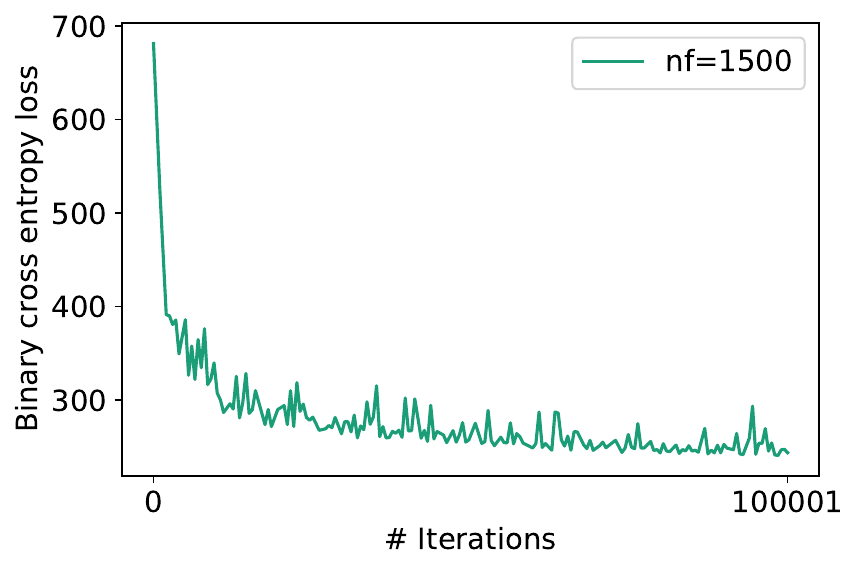}
  \caption{Verification error during training process for increasing number of feature points with full and diagonalized implementations.
  represented using solid and dashed lines respectively.
  Verification error in distributed diagonalized algorithm with $1500$ feature points.}
  \label{fig:error_full_diag}
\end{figure}

Fig.~\ref{fig:error_full_diag} compares the accuracy achieved with full covariance and diagonalized covariance 
estimates on varying number of feature points. 
For the same number of feature points, the full covariance updates are more accurate than the diagonalized ones.
The computational time with full covariance updates is an order of magnitude longer than diagonalized version. \
Therefore, we recommend that increasing the number of feature points over performing full covariance estimates 
for increasing predictive accuracy.

\begin{figure}[h]
  \centering
  \includegraphics[width=\linewidth]{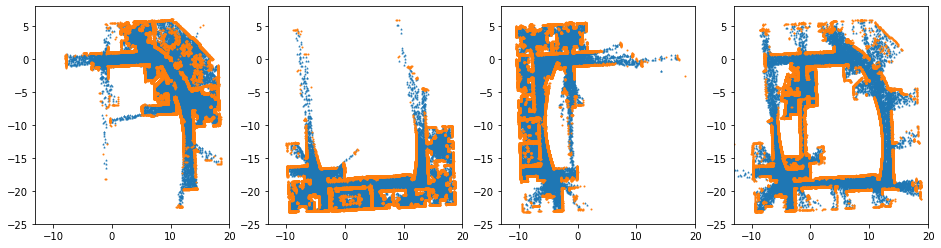}
  \includegraphics[width=0.43\linewidth, trim={1cm 0 0 0}]{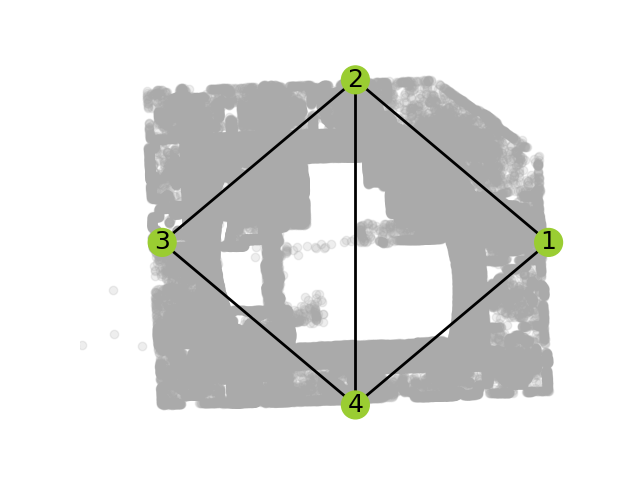}
  \includegraphics[width=0.55\linewidth]{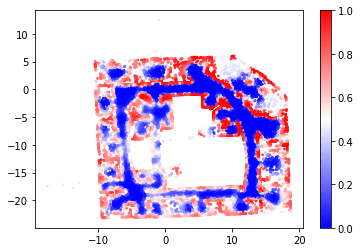}
  \caption{Training data distributed among $4$ agents sharing their inferences (top), 
  Communication network,
  Occupancy probability indicating free and occupied spaces 
  in blue and orange color respectively with a $1500$ feature point model.}
  \label{fig:dslam}
\end{figure}

As seen in Fig.~\ref{fig:dslam}, 
we distribute a reduced dataset with 290k (out of 380k) sequential points across four agents, 
such that only their combined dataset has the complete map information.
The agents communicate over a static connected graph in bottom-left of Fig.~\ref{fig:dslam}.
The $1500$ feature points and lengthscales $\gamma_2 = 0.5$ are selected at random 
from the test set as in the centralized setting,
and these points are common across the agents.
We achieve approximately $87\%$ predictive accuracy on the same test set.
Due to the presence of several agents, a quarter of iterations were sufficient to achieve 
this binary cross-entropy error as the centralized setting. 
The agents estimate similar mean values but their variances are lower for points close to the data collected.
% \NA{Provide a lot more details about the setup. 
% Where were the feature points, what was the observation model, 
% how long does it take to perform one iteration of the algorithm, 
% what was the communication graph, 
% plot classification error vs number of iterations from the perspective of agent $i$.}

\begin{figure}[h]
  \centering
  \includegraphics[width=\linewidth]{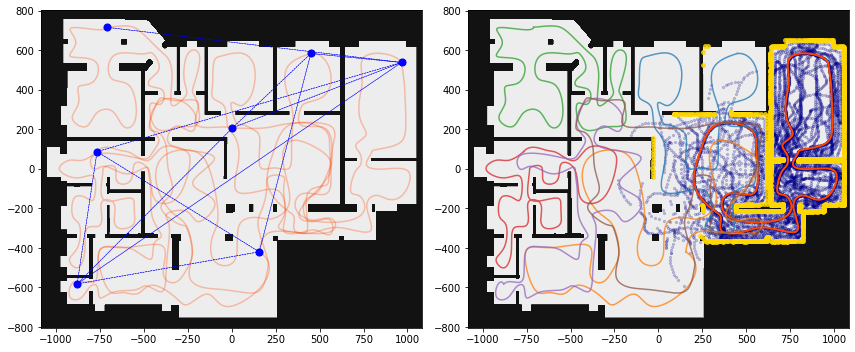}
  \includegraphics[width=0.49\linewidth]{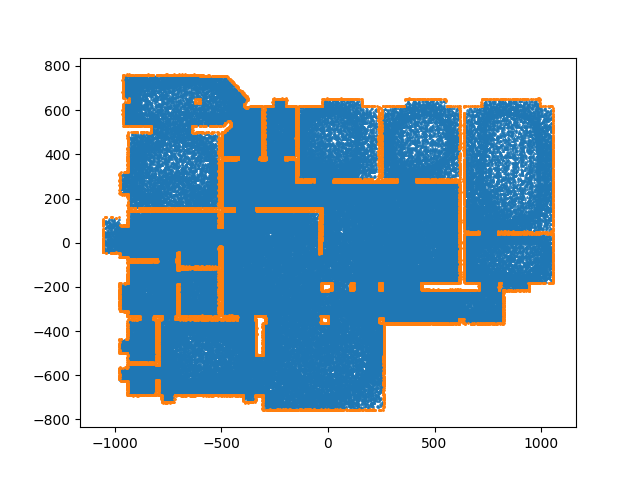}
  \includegraphics[width=0.49\linewidth]{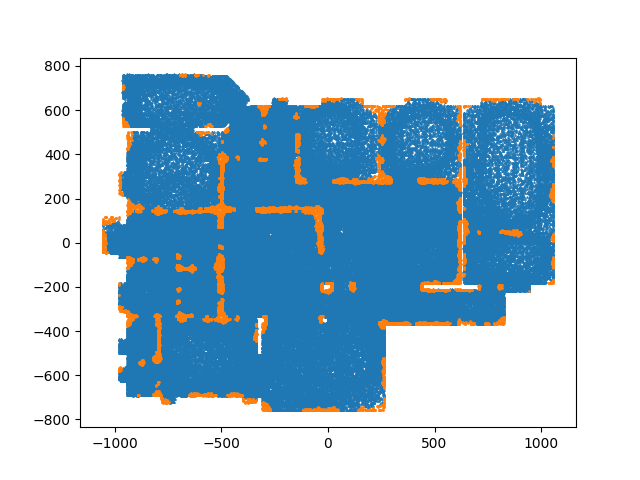}
  \includegraphics[width=0.49\linewidth]{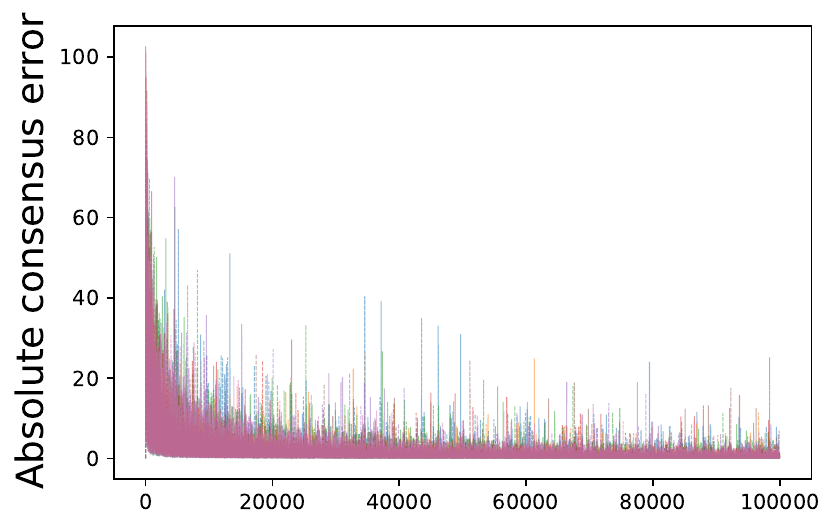}
  \includegraphics[width=0.49\linewidth]{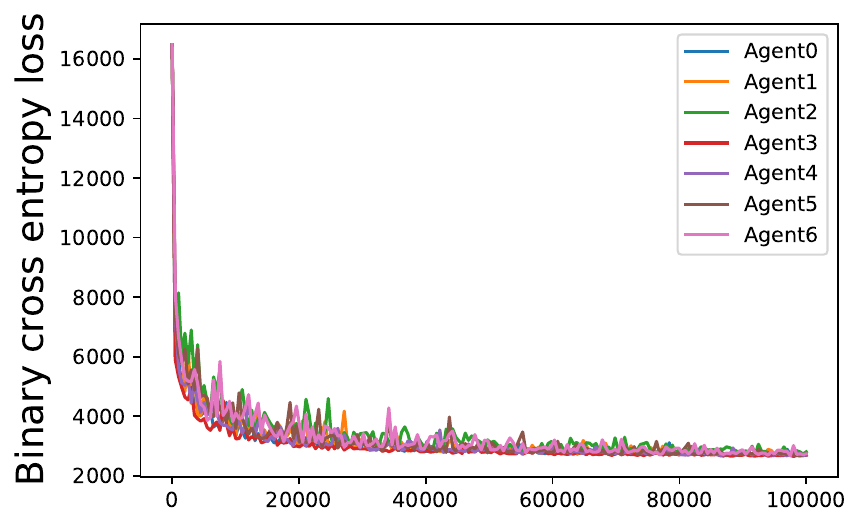}
  \caption{Data collected from the work in \cite{JY-JAV-MS:22}.
  Communication network laid over the trajectories corresponding to $7$ robots (top-left). 
  The collected LiDAR samples for first agent and remaining trajectories (top-right).
  Blue and yellow dots correspond to data indicating free and occupied spaces (middle).
  True and predicted point classes with a $1000$ feature point model. 
  Free and occupied spaces in blue and orange colors respectively.
  Consensus error summed over parameters for each agent (left-bottom) 
  and verification set error for each agent during training (bottom).}
  \label{fig:dslam_dinno}
\end{figure}

\paragraph*{DiNNO dataset \cite{JY-JAV-MS:22}} This dataset simulates LiDAR samples collected 
by multiple robots following independent trajectories with some overlap in observed environment.
In contrast to Intel dataset where we separated the data into four sets, 
here the robots have pre-determined trajectories with minimal overlap in indoor space. 
The LiDAR distance data is converted to five free and occupied points as shown at the top 
of Fig.~\ref{fig:dslam_dinno}. 
The training set consists of a third of the dataset, an-eleventh for test set and an-eightyeth for verification, 
chosen by slicing them along the trajectory.
Each of the seven robots has roughly $90k$ training points, with $175k$ points in the test set. 
This dataset is challenging due to the low number of occupied points $(10\%)$ in comparison to the ones in free space. 
Therefore, we choose $300$ feature points from the occupied space and remaining $700$ randomly.
Each kernel is defined with lengthscales $\gamma_2$ in $\{0.3, 3.\}$ depending on whether 
the data was chosen from occupied or free spaces respectively.
The reconstruction of the indoor space using the diagonal version of GVI is shown in Fig.~\ref{fig:dslam_dinno}.

The consensus error on the mean value of the parameters is computed as 
the deviation of the means $|\mu_{i,t}(\theta) - \frac{1}{n} \sum_{i=1}^{n} \mu_{i,t}(\theta)|$.
We can see that this error decreases with the number of iterations, implying that agents learn a common estimate.
During the training phase, prediction error is computed every $500$ iterations on the verification set with $23k$ instances.
The prediction error reaches a floor value over the $100k$ iterations for all agents.

\paragraph*{Successful training and deployment}
The theoretical derivation of DELBO assumes that independent observations at each agent.
In mapping data generated from robot trajectories, this assumption is not satisfied.
Therefore, we have used the idea of replay buffer to store data collected until a time 
and sample independently. 
While decomposing each distance measurement into points in free and occupied space, 
it is better to balance the points in each class while covering the entire space. 
We have maintained a $80-20$ ratio for the DiNNO dataset, 
more skewed than the Intel dataset.

Another key to building a good map is appropriate selection of feature points and 
lengthscales. The order of selected lengthscales should match the represented features.
For instance, the occupied spaces in the map should be represented with lengthscales
matching the expected obstacle width. 
In maps with several obstacle sizes, one could choose multiple kernels with varying 
lengthscales at the same feature points.
Greater density of feature points allow a detailed representation of geometric map 
features.
Selecting them from both occupied and free spaces allows better representation of 
each set. We selected $40\%$ of feature points in the occupied set to afford 
a better predictive resolution for DiNNO dataset.

%% file: appendix_simplification.tex
%%%%%%%%%%%%%%%%%%%%%%%%%%%%%%%%%%%%%%%%%%%%%%%%%%%%%%%%%%%%%%%%%%%%%%%%%%%%%%%%%%%%%%%%%%
\subsection{Gaussian variational inference}
\label{app:secgaussianalgo}
%%%%%%%%%%%%%%%%%%%%%%%%%%%%%%%%%%%%%%%%%%%%%%%%%%%%%%%%%%%%%%%%%%%%%%%%%%%%%%%%%%%%%%%%%%
\begin{proof}[Lemma~\ref{lemma:gvis}]
  \label{proof:gvi_sample}
  First, we discuss the derivation of the variational inference algorithm from the
  gradient descent steps in \cite{TDB-JRF-DJY:20}.
  We start by defining the objective function $\tau$ based on the known pdf $q_t$, 
  and follow up with its gradients,
  \begin{align}
    & \tau(\theta) = - \log \qz{}(z_{t+1}|\theta) - \log(q_t(\theta)), q_t(\theta) = \phi(\theta| \mu_t, \Sigma_t). \nonumber\\
    & \delta \mu = \mu_{t+1} - \mu_{t}
    = - \Omega_{t+1}^{-1} \underset{q_t}{\expect} \left[ \frac{\partial }{\partial \theta^{\top}} \tau(\theta) \right], \\
    & \Omega_{t+1} = \underset{q_t}{\expect} \left[ \frac{\partial }{\partial \theta^{\top} \partial \theta} \tau(\theta) \right]. \\
    & \frac{\partial }{\partial \theta^{\top}} \tau(\theta) = 
    - \frac{\partial}{\partial \theta^{\top}} [\log \qz{}(z_{t+1}|\theta)] + (\theta - \mu_t)^{\top} \Omega_t, \\
    & \underset{q_t}{\expect} \left[ \frac{\partial }{\partial \theta^{\top}} \tau(\theta)\right] 
    = - \underset{q_t}{\expect} \frac{\partial}{\partial \theta^{\top}} [\log \qz{}(z_{t+1}|\theta)]. \nonumber \\
    & \frac{\partial }{\partial \theta^{\top} \partial \theta} \tau(\theta) = 
    - \frac{\partial}{\partial \theta^{\top} \partial \theta} [\log \qz{}(z_{t+1}|\theta)] + \Omega_t, \\
    & \underset{q_t}{\expect} \left[ \frac{\partial }{\partial \theta^{\top}\partial \theta} \tau(\theta)\right] 
    = \Omega_t - \underset{q_t}{\expect} \frac{\partial}{\partial \theta^{\top}\partial \theta} [\log \qz{}(z_{t+1}|\theta)]. \nonumber 
  \end{align}
  The updated mean and information matrix are given as,
  \begin{align} 
    \label{eqn:update_grad}
    \begin{split}
      &\mu_{t+1} = \mu_t + \Omega_{t+1}^{-1} \underset{q_t}{\expect} \left[ \frac{\partial}{\partial \theta^{\top}} [\log \qz{}(z_{t+1}|\theta)] \right],\\
      & \Omega_{t+1} = \Omega_t - \underset{q_t}{\expect} \left[ \frac{\partial}{\partial \theta^{\top}\partial \theta} [\log \qz{}(z_{t+1}|\theta)] \right].  
    \end{split}
  \end{align}
This relates mean and covariance updates to the gradient and Hessian of the log-likelihood samples.
\end{proof}
% \paragraph*{Implicit: Diagonal form} Consider the ELBO term $V[q] = \underset{q(\theta)}{\expect} [\log \qz{}(z_{t+1}|\theta) - \log(q(\theta)) 
% + \log q_{t}(\theta)]$. Assume that the densities $q(\mu, D)$ and $q_{t}(\mu_t, D_t)$ 
% have diagonalized covariances given as vectors $D, D_t$.
% \begin{align*}
%   \KL(q || q_t) = \frac{1}{2} \sum_{k = 1}^l \left[  \log ( D_t[k]/D[k]) 
%   + D[k]/D_t[k] + 1/D_t[k](\mu_t[k] - \mu[k])^2 - 1 \right] \\
%   \frac{\partial}{\partial \mu[k]}\KL(q || q_t) = -\frac{1}{D_t[k]}(\mu_t[k] - \mu[k]),
%   \frac{\partial}{\partial D[k]}\KL(q || q_t) = \frac{1}{D_t[k]}-\frac{1}{D[k]}
% \end{align*}
% Computing $\frac{\partial}{\partial \mu} \underset{q(\theta)}{\expect} [\log \qz{}(z_{t+1}|\theta)] $,
% \begin{align*}
%   \frac{\partial}{\partial \mu} q(\theta) = - \frac{\partial}{\partial \theta} q(\theta) , 
%   \frac{\partial}{\partial D} q(\theta) = -1/2 (D^{-1} - (\theta - \mu)^2) q(\theta) = \frac{\partial^2}{\partial^2 \theta} q(\theta)
% \end{align*}
% \begin{align*}
%   \frac{\partial}{\partial \mu} \underset{q(\theta)}{\expect} [\log \qz{}(z_{t+1}|\theta)]
%   = \int \frac{\partial}{\partial \mu} q(\theta) \log \qz{}(z_{t+1}|\theta) d \theta 
%   = - \int \frac{\partial}{\partial \theta} q(\theta) \log \qz{}(z_{t+1}|\theta) d \theta 
% \end{align*}

%%%%%%%%%%%%%%%%%%%%%%%%%%%%%%%%%%%%%%%%%%%%%%%%%%%%%%%%%%%%%%%%%%%%%%%%%%%%%%%%%%%%%%%%%%
\subsection{Expectation of classification model with Gaussian density}
\label{app:expect}
%%%%%%%%%%%%%%%%%%%%%%%%%%%%%%%%%%%%%%%%%%%%%%%%%%%%%%%%%%%%%%%%%%%%%%%%%%%%%%%%%%%%%%%%%%
\begin{proof}[Expected gradient in Lemma~\ref{lemma:gradient}] 
  From Eqn.~\ref{eqn:loglikgrad_sigmoid}, the gradient of sigmoid function is, 
  $\nabla_\theta \log \qz{}(z| \theta) = (y - \sigma(\Phi_x^{\top} \theta)) \Phi_x^{\top}$.
  Its expected value with $q(\theta) \sim \calN(\mu, \Sigma)$ follows from the expectation of the term $\sigma(\Phi_x^{\top} \theta)$.
  For this computation, we recall the inverse probit function, or a cumulative distribution function 
  defined as $\Gamma(\theta) = \int_{\alpha = -\infty}^{\theta} \phi(\alpha)d\alpha$. 
  The cdf approximates the sigmoid function with the relationship $\sigma(\theta) = \Gamma(\xi \theta)$ 
  for $\xi = 0.61$ \cite{JD:17}. 
  To compute the approximation $\expect_{q(\theta)} [\Gamma(\xi \Phi_x^{\top} \theta)]$, 
  we substitute $u = \xi \Phi_x^{\top} \theta$ and express the cdf at $u$ 
  in terms of standard normal random variable $Z$ as $\Gamma(u) = \Pe(Z \leq U|U=u)$. Therefore,
  % Since $q(\theta)$ defines the expectation on $u$, 
  \begin{align*}
    \underset{q(\theta)}{\expect} [\Gamma(U)] = \underset{q(\theta)}{\expect} [\Pe(Z \leq U|U=u)] 
    %= \Pe(Z\leq u) 
    = \Pe(Z - U \leq 0).
  \end{align*}
  Since the variables $Z, U$ are jointly Gaussian, and $U$ is an affine transformation of $\Theta$, 
  their pdf can be expressed as $Z - U = \phi(\cdot| - \xi \Phi_x^{\top} \mu, 1 + \xi^2 \Phi_x^{\top} \Sigma \Phi_x)$,
  \begin{align}
    \Pe(Z - U \leq 0) = \Gamma \left( 
      \frac{\xi \Phi_x^{\top} \mu}{\sqrt{1 + \xi^2 \Phi_x^{\top} \Sigma \Phi_x}}\right)
  \end{align}  
  With $\beta = 1 + \xi^2 \Phi_x^{\top} \Sigma \Phi_x$, the approximate expected value of 
  the sigmoid function in the gradient defined in Eqn.~\ref{eqn:loglikgrad_sigmoid} is,
  \begin{align*}
    & \underset{q_{t}(\theta)}{\expect} [\sigma(\Phi_x^{\top}\theta)] 
    \approx \int \Gamma(\xi \Phi_x^{\top}\theta) q_{t}(\theta) d \theta \nonumber 
     = \Gamma \left( \frac{\xi \Phi_x^{\top} \mu_{t}}{\sqrt{\beta}} \right).
  \end{align*}
  Thus, the expected gradient of the log-likelihood is,
  \begin{align*}
    \underset{q_t}{\expect} [(y - \sigma(\Phi_x^{\top} \theta)) \Phi_x^{\top}]
    = \left(y - \Gamma \left( \xi \Phi_x^{\top} \mu_t \middle/
    \sqrt{\beta}\right)\right) \Phi_x^{\top}.
  \end{align*}
\end{proof}

\begin{proof}[Expected Hessian in Lemma~\ref{lemma:gradient}] 
  To find a tractable analytical expression for the new covariance matrix $\Omega_{t+1}^{-1}$,
We start by computing the expectation from Eqn.~\ref{eqn:loglikhess_sigmoid}, 
\begin{align*}
  & \expect_{q_{t}} [\phi (\xi \Phi_x^{\top} \theta| 0, 1)] 
  = \int q_{t}(\theta) \phi (\xi \Phi_x^{\top} \theta| 0, 1) d \theta ,\\
  & = \sqrt{\frac{\det(\Omega_t)}{(2 \pi)^{d_{\theta}+1}}} \int_{\theta} 
  \exp\left(-\frac{1}{2} (\theta - \mu_t)^{\top}\Omega_t (\theta - \mu_t)\right) \\
  & \qquad \exp\left(-\frac{1}{2}\xi^2 \theta^{\top} \Phi_x \Phi_x^{\top} \theta\right) d \theta ,\\
  & = \sqrt{\frac{\det(\Omega_t)}{(2 \pi)^{d_{\theta}+1}}} \int_{\theta} 
  \exp\left(-\frac{1}{2} [\theta^{\top} (\Omega_t + \xi^2 \Phi_x \Phi_x^{\top}) \theta \right. \\
  & \qquad \left.- 2 \theta^{\top} \Omega_t \mu_t + \mu_t^{\top} \Omega_t \mu_t  ] \right) d \theta.
\end{align*}

Define $A = \Omega_t + \xi^2 \Phi_x \Phi_x^{\top}, b = \Omega_t \mu_t, 
c = \mu_t^{\top} \Omega_t \mu_t$ in the argument of quadratic exponential
to proceed with sum of squares technique, 
\begin{align*}
  & \expect_{q_{t}} [\phi (\xi \Phi_x^{\top} \theta| 0, 1)] \\
  % & = \sqrt{\frac{\det(\Omega_t)}{(2 \pi)^{d_{\theta}+1}}} \int_{\theta} 
  % \exp\left(-\frac{1}{2} [(\theta- A^{-1}b)^{\top} A (\theta- A^{-1}b) 
  % - b^{\top}A^{-1}b + \mu_t^{\top} \Omega_t \mu_t  ] \right) d \theta \\
  & = \sqrt{\frac{\det(\Omega_t)}{(2 \pi)^{d_{\theta}+1}}} 
  \exp\left(-\frac{1}{2} [ - b^{\top}A^{-1}b + \mu_t^{\top} \Omega_t \mu_t  ] \right) \\
  & \quad \int_{\theta} 
  \exp\left(-\frac{1}{2} [(\theta- A^{-1}b)^{\top} A (\theta- A^{-1}b)  ] \right) d \theta ,  \\
  % & = \sqrt{\frac{\det(\Omega_t)}{(2 \pi)^{d_{\theta}+1}}}
  % \exp\left(-\frac{1}{2} [ - \mu_t^{\top} \Omega_t^{\top}(\Omega_t + \xi^2 \Phi_x \Phi_x^{\top})^{-1}
  % \Omega_t \mu_t + \mu_t^{\top} \Omega_t \mu_t  ] \right)
  % (2 \pi)^{\frac{d_{\theta}}{2}} \det(A)^{-1/2} \\
  & = \sqrt{\frac{\det(\Omega_t)}{2 \pi \det(\Omega_t + \xi^2 \Phi_x \Phi_x^{\top})}} \\
  & \;\exp\left(-\frac{1}{2} [ - \mu_t^{\top} \Omega_t^{\top}(\Omega_t + \xi^2 \Phi_x \Phi_x^{\top})^{-1}
  \Omega_t \mu_t + \mu_t^{\top} \Omega_t \mu_t  ] \right).
\end{align*}

Since computing the determinant and the inverse in the previous formula is expensive,
we employ the matrix determinant lemma stating that $\det(\Omega_{t} + \xi^2\Phi_x \Phi_x^{\top}) = 
  (1+\xi^2\Phi_x^{\top} \Omega_{t}^{-1} \Phi_x)\det(\Omega_{t})$.
  \begin{align*}
    \sqrt{\frac{\det(\Omega_t)}{2 \pi \det(\Omega_t + \xi^2 \Phi_x \Phi_x^{\top})}}
    = \sqrt{\frac{1}{2 \pi (1+\xi^2\Phi_x^{\top} \Omega_{t}^{-1} \Phi_x)}}.
  \end{align*}
  The inverse of the dense matrix $(\Omega_{t-1} + \xi^2\Phi_x \Phi_x^{\top})^{-1}$ can be simplified using Woodbury's formula \cite{MAW:50} 
  such that we use the precomputed covariance matrix $\Omega_{t-1}^{-1}$ along with a scalar inverse. In batch settings,
  this inverse is over low dimensions in comparison to number of feature points $l$.
  \begin{align*}
    & (\Omega_{t} + \xi^2\Phi_x \Phi_x^{\top})^{-1} \\
    & = \Omega_{t}^{-1} - \xi^2 \Omega_{t}^{-1}\Phi_x( 1 + \xi^2\Phi_x^{\top} 
    \Omega_{t}^{-1} \Phi_x)^{-1}\Phi_x^{\top} \Omega_{t}^{-1}.
  \end{align*}
  Substituting $\beta = 1 + \xi^2 \Phi_x^{\top} \Omega_{t}^{-1} \Phi_x$, 
  the expected second order derivative is thus simplified as,
  \begin{align}
    \expect_{q_{t}} &[\nabla_{\theta}^2 \log p (z_t|\theta)] \nonumber \\
    & = -  \sqrt{\frac{\xi^2}{2 \pi \beta}}
    \exp\left(-\frac{1}{2} [\frac{\xi^2}{\beta} \mu_t^{\top} \Phi_x \Phi_x^{\top} \mu_t] \right)\Phi_x\Phi_x^{\top}, \nonumber\\
    \Omega_{t+1} & = \Omega_{t} + \gamma \Phi_x \Phi_x^{\top}.
\end{align}
Thus, we have a linear update for the information matrix.
\end{proof}

\begin{proof}[Lemma~\ref{lemma:dgvi_gaussian}]
  The mean and covariance updates at any agent $i$ follow from gradient and Hessians of the likelihood 
  w.r.t. the mixed pdf $q_{i,t}^g = \prod_j q_{j,t}^{A_{ij}}$.
  A computationally cheap method to compute the inverse of information matrix $\Omega_{t+1}$ 
  in the expression of the next mean value in Eqn.~\ref{eqn:update_grad} is derived from the matrix inversion lemma \cite{MAW:50} as,
  \begin{align*}
    \Omega_{t+1}^{-1} & = \Omega_{t}^{-1} - \gamma \Omega_{t}^{-1}\Phi_x( I + \gamma\Phi_x^{\top} 
    \Omega_{t}^{-1} \Phi_x)^{-1}\Phi_x^{\top} \Omega_{t}^{-1}.
    % \Sigma_{t+1} & = \Sigma_t - \gamma \Sigma_t\Phi_x( I + \gamma\Phi_x^{\top} 
    % \Sigma_t \Phi_x)^{-1}\Phi_x^{\top} \Sigma_t
  \end{align*}  
  In a single agent setting, this avoids performing any matrix inverse after the initial step.
\end{proof}

%%%%%%%%%%%%%%%%%%%%%%%%%%%%%%%%%%%%%%%%%%%%%%%%%%%%%%%%%%%%%%%%%%%%%%%%%%%%%%%%%%%%%%%%%%
\subsection{Diagonal Gaussian derivation}
\label{app:diag_G}
%%%%%%%%%%%%%%%%%%%%%%%%%%%%%%%%%%%%%%%%%%%%%%%%%%%%%%%%%%%%%%%%%%%%%%%%%%%%%%%%%%%%%%%%%%
\begin{proof}[Proof for Lemma~\ref{lemma:diag_dgvi}]
  We follow the approach in \cite{TDB-JRF-DJY:20} but with additional diagonalized 
  approximation of the second-order Taylor expansion and elementwise derivatives 
  over the diagonal terms in the information matrix. 
  Assume that the densities $q(\mu, D)$ and $q_{t}(\mu_t, D_t)$ 
  have diagonalized information matrices with diagonal vectors $\Delta, \Delta_t$ whose 
  $i$-th elements are $\Delta_i, \Delta_{i,t}$. With $\tau(\theta) = - \log q(z|\theta) - \log p(\theta)$,
  the variational objective is, 
  \begin{align*}
    & V(q) = \underset{q}{\expect} [\tau(\theta) + \log q(\theta)] = \frac{1}{2} \sum_{i=1}^l \log \Delta_i \\
    & + \int \tau(\theta) \prod_{i=1}^l (\frac{2 \pi}{\Delta_i})^{-l/2} \exp \left( - \frac{1}{2} 
    \sum_{i=1}^l \Delta_i (\theta_i - \mu_i)^2\right) d \theta.
  \end{align*}

  The derivatives of ELBO w.r.t. the mean and information matrix are given as,
  \begin{align*}
    & \frac{\partial}{\partial \mu^{\top}} V(q) = D \int \tau(\theta) q(\theta) (\theta - \mu) d \theta
    = D \, \underset{q}{\expect} [\tau(\theta) (\theta - \mu)], \\
    & \frac{\partial^2}{\partial \mu^{\top}\partial \mu} V(q) 
    = D \, \underset{q}{\expect} [\tau(\theta) (\theta - \mu) (\theta - \mu)^{\top}] D 
    - D \underset{q}{\expect} [\tau(\theta)],
  \end{align*}
  \begin{align*}
    \frac{\partial  V(q)}{\partial D} & = \frac{1}{2} 
    \underset{q}{\expect} [\tau(\theta)(D^{-1} - (\theta - \mu) (\theta - \mu)^{\top})]
    + \frac{1}{2} D^{-1}.
  \end{align*}

  The derivative of the objective w.r.t. scalar terms in the mean and information matrix diagonal are,
  \begin{align*}
    \frac{\partial}{\partial \mu_i} V(q)
    = & \Delta_i \int q(\theta) \tau(\theta) (\theta_i - \mu_i)d\theta ,\\
    \frac{\partial^2}{\partial \mu_i^{2}} V(q)  
    = & \int q(\theta) \tau(\theta) (\Delta_i^2 (\theta_i - \mu_i)^2 - \Delta_i) d\theta ,\\
    \frac{\partial}{\partial \Delta_i} V(q) 
    = & -\frac{1}{2} \int q(\theta) \tau(\theta) (\theta_i - \mu_i)^2 d\theta \\
    & +\frac{1}{2\Delta_i} \left( \int q(\theta) \tau(\theta) d\theta + 1 \right).
  \end{align*}
  The double derivative w.r.t. the mean is related to the one from information matrix as,
  \begin{align*}
    \frac{\partial^2}{\partial \mu_i^{2}} V(q) = 
    -2 \Delta_i^2 \frac{\partial}{\partial \Delta_i} V(q) + \Delta_i .
  \end{align*}

  Since $\frac{\partial}{\partial \Delta_i} V(q) = 0$ for all $i$ at the local optimum, 
  we can claim that,
  \begin{align*}
    \Delta_{i, t+1} = \left. \frac{\partial^2}{\partial \mu_i^{2}} V(q) \right\vert_{q_t}, 
    D_{t+1} = \diag{\frac{\partial^2}{\partial \mu^{\top}\partial \mu} V(q)} . % \\
    % \implies \Delta_{i, t+1} - \Delta_{i, t} = - 2 \Delta_i^2 \frac{\partial}{\partial \Delta_i} V(q).
  \end{align*}

  As shown in \cite{TDB-JRF-DJY:20}, we can approximate the value of function $V$ in terms of 
  vector differentials on mean $\delta \mu = \mu_{t+1} - \mu_t$ 
  and information diagonal $\delta \Delta = \Delta_{t+1} - \Delta_t$.
  \begin{align*}
    & V(q_{t+1}) \approx V(q_t) +  \left. \frac{\partial}{\partial \mu} V(q)\right\vert_{q_t} \delta\mu 
    + \left. \frac{\partial}{\partial \Delta} V(q)\right\vert_{q_t} \delta\Delta \\
    & + \frac{1}{2} \delta \mu^{\top} \left.\frac{\partial^2}{\partial \mu^{\top} \partial \mu} V(q)\right\vert_{q_t} \delta \mu
    , \tag{Taylor expansion}\\
    & \approx V(q_t) +  \left. \frac{\partial}{\partial \mu} V(q)\right\vert_{q_t} \delta\mu 
    + \left. \frac{\partial}{\partial \Delta} V(q)\right\vert_{q_t} \delta\Delta \\
    & + \frac{1}{2} \delta \mu^{\top} \diag{\left.\frac{\partial^2}{\partial \mu^{\top} \partial \mu} V(q)\right\vert_{q_t}} \delta \mu .
    \tag{Diagonal Hessian}
  \end{align*}
  The diagonal approximation of the Hessian matrix holds if the underlying log-likelihood model $\log \qz{}(z | \theta)$
  is almost linear in terms of parameters $\theta$.
  Since the objective is locally quadratic in $\delta \mu$, we can set the derivative w.r.t. $\delta \mu$
  to zero, leading to a linear system of the form,
  \begin{align*}
    &\mathrm{diag}\left( \frac{\partial^2}{\partial \mu^{\top}\partial \mu} V(q) \right) \delta\mu = \frac{\partial}{\partial \mu^{\top}} V(q), \\ 
    &\delta\mu = D_{t+1}^{-1} \left( \frac{\partial}{\partial \mu^{\top}} V(q) \right).
  \end{align*}

  Using Stein's lemma \cite{CMS:81},
  \begin{align*}
    & \underset{q}{\expect} [(\theta - \mu) \tau(\theta)] \equiv D^{-1} \underset{q}{\expect} 
    \left[ \frac{\partial \tau(\theta)}{\partial \theta^{\top}} \right] = D^{-1} \frac{\partial}{\partial \mu^{\top}} V(q), \\
    & \underset{q}{\expect} [\tau(\theta) (\theta - \mu) (\theta - \mu)^{\top}]  \\
    & \quad \equiv D^{-1} \underset{q}{\expect} \left[\frac{\partial^2 \tau(\theta)}{\partial \theta^{\top} \partial \theta } \right] D^{-1} 
    + D^{-1} \underset{q}{\expect} \left[\tau(\theta) \right], \\
    & \frac{\partial^2}{\partial \mu^{\top}\partial \mu} V(q) = 
    \underset{q}{\expect} \left[\frac{\partial^2 \tau(\theta)}{\partial \theta^{\top} \partial \theta } \right], 
    \frac{\partial^2}{\partial \mu_i^{2}} V(q) = 
    \underset{q}{\expect} \left[\frac{\partial^2 \tau(\theta)}{\partial \theta_i^{2} } \right].
  \end{align*}
  Therefore, the udpate rules for mean and diagonal information matrix are,
  \begin{align*}
    & D_{t+1} = \mathrm{diag}\left( \underset{q}{\expect} \left[\frac{\partial^2 \tau(\theta)}{\partial \theta^{\top} \partial \theta } \right] \right),\\
    & \mu_{t+1} - \mu_t = D_{t+1}^{-1} \underset{q}{\expect} \left[ \frac{\partial \tau(\theta)}{\partial \theta^{\top}} \right].
  \end{align*}
  Using the simplification in Appendix~\ref{app:secgaussianalgo} followed by 
  expectation of the classification model in Appendix~\ref{app:expect} and diagonalized $D_t$, 
  we obtain the updates,
  \begin{align*}
    D_{t+1} &= \mathrm{diag}\left(D_t + \gamma \Phi_x \Phi_x^{\top} \right) = D_t + \gamma \mathrm{diag}\left(\Phi_x \Phi_x^{\top} \right),\\
    \mu_{t+1} & - \mu_t =  D_{t+1}^{-1} \underset{q_t}{\expect} \left[ \frac{\partial}{\partial \theta^{\top}} [\log \qz{}(z_{t+1}|\theta)] \right],\\
    & \qquad = D_{t+1}^{-1} \expect_{q_{t}} [(y - \sigma(\Phi_x^{\top} \theta)) \Phi_x^{\top}], \\
    & \qquad = D_{t+1}^{-1} (y - \Gamma \left( \xi \Phi_x^{\top} \mu \big/ \sqrt{\beta} \right)) \Phi_x^{\top}.
  \end{align*}
  Here, $\gamma = \sqrt{\frac{\xi^2}{2 \pi \beta}}
  \exp\left(-\frac{1}{2} [\frac{\xi^2}{\beta} \mu_t^{\top} \Phi_x \Phi_x^{\top} \mu_t] \right)$,
  with $\beta = 1 + \xi^2 \Phi_x^{\top} D_t^{-1} \Phi_x$.
\end{proof}

%%%%%%%%%%%%%%%%%%%%%%%%%%%%%%%%%%%%%%%%%%%%%%%%%%%%%%%%%%%%%%%%%%%%%%%%%%%%%%%%%%%%%%%%%%
\subsection{Distributed regression in Gaussian models}
\label{app:regression}
%%%%%%%%%%%%%%%%%%%%%%%%%%%%%%%%%%%%%%%%%%%%%%%%%%%%%%%%%%%%%%%%%%%%%%%%%%%%%%%%%%%%%%%%%%
Let the linear regression model with parameters $\theta$ describe the relationship between 
input output pairs $z =(x, y)$ at agent $i$ be specified as the likelihood 
$\qz{i}(z|\theta) \propto \exp(-0.5(y - \Phi_{x}^{\top} \theta)^{\top}S_i(y - \Phi_{x}^{\top} \theta))$, where $S_i$ is positive definite.
Following the steps for the classification problem, the log likelihood gradient and Hessian terms are,
\begin{align*}  
  & \nabla_{\theta} \log p(z_i|\theta) = \Phi_x S_i(y -  \Phi_x^{\top} \theta), \\
  & \nabla_{\theta}^2 \log p(z_i|\theta) = -\Phi_x S_i \Phi_x^{\top}.
\end{align*}
The mixed Gaussian pdf $q_{i, t}^g = \calN(\theta|\mu_{i,t}^g, \Sigma_{i,t}^{g})$ 
for regression follows from Lemma~\ref{lemma:dgvi_gaussian} with $\Sigma_{i,t}^{g} = (\Omega_{i,t}^{g})^{-1}$,
\begin{align*}
  \Omega_{i,t}^g = \sum_{j \in \nodes} A_{ij} \Omega_{j, t}, 
   \mu^g_{i, t} = (\Omega_{i,t}^g)^{-1} \sum_{j \in \nodes} A_{ij} \Omega_{j, t}\mu_{j, t}.
\end{align*}

Then, we can follow~\eqref{eqn:update_grad} and Woodbury's matrix inversion lemma \cite{MAW:50}
w.r.t. $q_{i, t}^g$, 
\begin{align*}
    & \Omega_{i,t+1} = \Omega_t - \underset{q_{i, t}^g}{\expect} [\nabla_{\theta}^2 \log \qz{i}(z|\theta)] 
    = \Omega_{i,t}^g + \Phi_x S_i \Phi_x^{\top},\\
    & \Omega_{i,t+1}^{-1} = \Sigma_{i,t}^g - \Sigma_{i,t}^g \Phi_x 
    (S_i^{-1} + \Phi_x^{\top} \Sigma_{i,t}^g \Phi_x)^{-1} \Phi_x^{\top} \Sigma_{i,t}^g, \nonumber \\
    & \mu_{i,t+1} = \mu_{i,t}^g + (\Omega_{i,t+1})^{-1} (\Phi_x S_i^{\top} y - \Phi_x S_i \Phi_x^{\top} \mu_{i,t}^g) .
\end{align*}
Thus, we have distributed probabilistic updates on the parameters of the linear regression model.
%

%% file: root.bbl
\begin{thebibliography}{10}

\bibitem{MA-JG-EX-AR:20}
M.~Al-Shedivat, J.~Gillenwater, E.~Xing, and A.~Rostamizadeh.
\newblock Federated learning via posterior averaging: A new perspective and
  practical algorithms.
\newblock {\em arXiv preprint arXiv:2010.05273}, 2020.

\bibitem{TDB-JRF-DJY:20}
T.~D. Barfoot, J.~R. Forbes, and D.~J. Yoon.
\newblock Exactly sparse {G}aussian variational inference with application to
  derivative-free batch nonlinear state estimation.
\newblock {\em Int. J. Rob. Res.}, 39(13):1473--1502, 2020.

\bibitem{AB-SE-JW:05}
A.~Bordes, S.~Ertekin, J.~Weston, L.~Botton, and N.~Cristianini.
\newblock Fast kernel classifiers with online and active learning.
\newblock {\em J. Mach. Learn. Res.}, 6(9), 2005.

\bibitem{TDB-CVN-SS-RET:18}
T.~D. Bui, C.~V. Nguyen, S.~Swaroop, and R.~E. Turner.
\newblock Partitioned variational inference: A unified framework encompassing
  federated and continual learning.
\newblock {\em arXiv preprint arXiv:1811.11206}, 2018.

\bibitem{FB-JC-SM:09}
F.~Bullo, J.~Cort\'es, and S.~Mart{\'\i}nez.
\newblock {\em Distributed Control of Robotic Networks}.
\newblock Applied Mathematics Series. Princeton University Press, 2009.
\newblock Electronically available at http://coordinationbook.info.

\bibitem{JC:21}
J.~Cadena, P.~Ray, H.~Chen, B.~Soper, D.~Rajan, A.~Yen, and R.~Goldhahn.
\newblock Stochastic gradient-based distributed {B}ayesian estimation in
  cooperative sensor networks.
\newblock {\em IEEE Trans. Signal Process.}, 69:1713--1724, 2021.

\bibitem{XC-TB:23}
X.~Cao, T.~Ba{\c{s}}ar, S.~Diggavi, Y.~C. Eldar, K.~B. Letaief, H.~V. Poor, and
  J.~Zhang.
\newblock Communication-efficient distributed learning: An overview.
\newblock {\em IEEE J. Sel. Areas Commun.}, 2023.

\bibitem{HD-YZ-JL:13}
H.~Dai, Y.~Zhang, and J.~Liu.
\newblock Structured variational methods for distributed inference in networked
  systems: Design and analysis.
\newblock {\em IEEE Trans. Signal Process.}, 61(15):3827--3839, 2013.

\bibitem{JD:17}
J.~Daunizeau.
\newblock Semi-analytical approximations to statistical moments of sigmoid and
  softmax mappings of normal variables.
\newblock {\em arXiv preprint arXiv:1703.00091}, 2017.

\bibitem{TD-MY-NA:22}
T.~Duong, M.~Yip, and N.~Atanasov.
\newblock Autonomous navigation in unknown environments with sparse {B}ayesian
  kernel-based occupancy mapping.
\newblock {\em IEEE Trans. Robot.}, 38(6):3694--3712, 2022.

\bibitem{DF:97}
D.~Fink.
\newblock A compendium of conjugate priors.
\newblock Technical report, 1997.

\bibitem{ZG-MB:00}
Z.~Ghahramani and M.~Beal.
\newblock Propagation algorithms for variational {B}ayesian learning.
\newblock {\em Adv. Neural Inf. Process Syst.}, 13, 2000.

\bibitem{SG-JP:17}
S.~Gultekin and J.~Paisley.
\newblock Nonlinear {K}alman filtering with divergence minimization.
\newblock {\em IEEE Trans. Signal Process.}, 65(23):6319--6331, 2017.

\bibitem{MH-DB-CW-JP:13}
M.~D. Hoffman, D.~M. Blei, C.~Wang, and J.~Paisley.
\newblock Stochastic variational inference.
\newblock {\em J. Mach. Learn. Res.}, 2013.

\bibitem{AH-NR:03}
A.~Howard and N.~Roy.
\newblock The robotics data set repository (radish), 2003.

\bibitem{JH-CL:15}
J.~Hua and C.~Li.
\newblock Distributed variational {B}ayesian algorithms over sensor networks.
\newblock {\em IEEE Trans. Signal Process.}, 64(3):783--798, 2015.

\bibitem{TSJ-MIJ:00}
T.~S. Jaakkola and M.~I. Jordan.
\newblock {B}ayesian parameter estimation via variational methods.
\newblock {\em Stat. Comput.}, 10(1):25--37, 2000.

\bibitem{MIJ-ZG-:99}
M.~I. Jordan, Z.~Ghahramani, T.~S. Jaakkola, and L.~K. Saul.
\newblock An introduction to variational methods for graphical models.
\newblock {\em Machine learning}, 37:183--233, 1999.

\bibitem{PK:21}
P.~Kairouz, H.~B. McMahan, B.~Avent, A.~Bellet, M.~Bennis, A.~N. Bhagoji,
  K.~Bonawitz, Z.~Charles, G.~Cormode, R.~Cummings, et~al.
\newblock Advances and open problems in federated learning.
\newblock {\em Found. Trends Mach. Learn.}, 14(1--2):1--210, 2021.

\bibitem{DPK-MW:19}
D.~P. Kingma, M.~Welling, et~al.
\newblock An introduction to variational autoencoders.
\newblock {\em Found. Trends Mach. Learn.}, 12(4):307--392, 2019.

\bibitem{JK-JJ-TD:22}
J.~Knoblauch, J.~Jewson, and T.~Damoulas.
\newblock An optimization-centric view on {B}ayes' rule: Reviewing and
  generalizing variational inference.
\newblock {\em J. Mach. Learn. Res.}, 23(132):1--109, 2022.

\bibitem{ML-SB-FB:22}
M.~Lambert, S.~Bonnabel, and F.~Bach.
\newblock The recursive variational {G}aussian approximation ({R-VGA}).
\newblock {\em Stat. Comput.}, 32(1):10, 2022.

\bibitem{PSL:86}
P.~S. Laplace.
\newblock Memoir on the probability of the causes of events.
\newblock {\em Statistical science}, 1(3):364--378, 1986.

\bibitem{XL:17}
X.~Lian, C.~Zhang, H.~Zhang, C.-J. Hsieh, W.~Zhang, and J.~Liu.
\newblock Can decentralized algorithms outperform centralized algorithms? a
  case study for decentralized parallel stochastic gradient descent.
\newblock {\em Adv. Neural Inf. Process Syst.}, 30, 2017.

\bibitem{JL:21}
J.~Luttinen.
\newblock {B}ayesian python: {B}ayesian inference tools for python, 2021.

\bibitem{SM-MH-DB:17}
S.~Mandt, M.~D. Hoffman, and D.~M. Blei.
\newblock Stochastic gradient descent as approximate {B}ayesian inference.
\newblock {\em J. Mach. Learn. Res.}, 18:1--35, 2017.

\bibitem{MAW:50}
A.~W. Max.
\newblock Inverting modified matrices.
\newblock In {\em Memorandum Rept. 42, Statistical Research Group}, page~4.
  Princeton Univ., 1950.

\bibitem{BM:17}
B.~McMahan, E.~Moore, D.~Ramage, S.~Hampson, and B.~A.~y. Arcas.
\newblock {Communication-Efficient Learning of Deep Networks from Decentralized
  Data}.
\newblock In A.~Singh and J.~Zhu, editors, {\em Proc. of the 20th Int. Conf. on
  Artificial Intelligence and Statistics}, volume~54, pages 1273--1282. PMLR,
  20--22 Apr 2017.

\bibitem{AN-AO-CAA:17}
A.~Nedi{\'c}, A.~Olshevsky, and C.~A. Uribe.
\newblock Fast convergence rates for distributed non-{B}ayesian learning.
\newblock {\em IEEE Trans. Autom. Control}, 62(11):5538--5553, 2017.

\bibitem{HN-CER:08}
H.~Nickisch and C.~E. Rasmussen.
\newblock Approximations for binary {G}aussian process classification.
\newblock {\em J. Mach. Learn. Res.}, 9(Oct):2035--2078, 2008.

\bibitem{VO-DN-MS:18}
V.~M.-H. Ong, D.~J. Nott, and M.~S. Smith.
\newblock {G}aussian variational approximation with a factor covariance
  structure.
\newblock {\em J. Comput. Graph. Stat.}, 27(3):465--478, 2018.

\bibitem{PP-NA-SM:19-cdc}
P.~Paritosh, N.~Atanasov, and S.~Mart{\'\i}nez.
\newblock Hypothesis assignment and partial likelihood averaging for
  cooperative estimation.
\newblock In {\em {IEEE} Int. Conf. on Decision and Control}, pages 7850--7856,
  Nice, France, December 2019.

\bibitem{PP-NA-SM:22}
P.~Paritosh, N.~Atanasov, and S.~Martinez.
\newblock Distributed {B}ayesian estimation of continuous variables over
  time-varying directed networks.
\newblock {\em IEEE Control Syst. Lett.}, 6:2545--2550, 2022.

\bibitem{RR-SG-DB:14}
R.~Ranganath, S.~Gerrish, and D.~Blei.
\newblock Black box variational inference.
\newblock In {\em Int. Conf. on Artificial Intelligence and Statistics}, pages
  814--822. PMLR, 2014.

\bibitem{DJR-SM-DW:14}
D.~J. Rezende, S.~Mohamed, and D.~Wierstra.
\newblock Stochastic backpropagation and approximate inference in deep
  generative models.
\newblock In {\em Proc. Int. Conf. Mach. Learn.}, pages 1278--1286. PMLR, 2014.

\bibitem{TS-AG:20}
T.~Shankar and A.~Gupta.
\newblock Learning robot skills with temporal variational inference.
\newblock In {\em Proc. Int. Conf. Mach. Learn.}, pages 8624--8633. PMLR, 2020.

\bibitem{RS-PK:67}
R.~Sinkhorn and P.~Knopp.
\newblock Concerning nonnegative matrices and doubly stochastic matrices.
\newblock {\em Pac. J. Math.}, 21(2):343--348, 1967.

\bibitem{VS-AQ:08}
V.~Smidl and A.~Quinn.
\newblock Variational {B}ayesian filtering.
\newblock {\em IEEE Trans. Signal Process.}, 56(10):5020--5030, 2008.

\bibitem{CMS:81}
C.~M. Stein.
\newblock Estimation of the mean of a multivariate normal distribution.
\newblock {\em The Annals of Statistics}, pages 1135--1151, 1981.

\bibitem{ST-DL-BW:22}
T.~Sun, D.~Li, and B.~Wang.
\newblock Decentralized federated averaging.
\newblock {\em IEEE Trans. Pattern Anal. Mach. Intell.}, 2022.

\bibitem{CAU-AO-AN:22}
C.~A. Uribe, A.~Olshevsky, and A.~Nedi{\'c}.
\newblock Nonasymptotic concentration rates in cooperative learning--part i:
  Variational non-bayesian social learning.
\newblock {\em IEEE Trans. Control Netw. Syst.}, 9(3):1128--1140, 2022.

\bibitem{XW-AL-TJ-FK:22}
X.~Wang, A.~Lalitha, T.~Javidi, and F.~Koushanfar.
\newblock Peer-to-peer variational federated learning over arbitrary graphs.
\newblock {\em IEEE J. Sel. Areas Inf. Theory}, 2022.

\bibitem{JW-CB-TJ:05}
J.~Winn, C.~M. Bishop, and T.~Jaakkola.
\newblock Variational message passing.
\newblock {\em J. Mach. Learn. Res.}, 6(4), 2005.

\bibitem{YT:19}
T.~Yang, X.~Yi, J.~Wu, Y.~Yuan, D.~Wu, Z.~Meng, Y.~Hong, H.~Wang, Z.~Lin, and
  K.~H. Johansson.
\newblock A survey of distributed optimization.
\newblock {\em Annu. Rev. Control}, 47:278--305, 2019.

\bibitem{JY-JAV-MS:22}
J.~Yu, J.~A. Vincent, and M.~Schwager.
\newblock {DiNNO}: Distributed neural network optimization for multi-robot
  collaborative learning.
\newblock {\em IEEE Robot. Autom. Lett.}, 7(2):1896--1903, 2022.

\bibitem{CZ-JB-HK:18}
C.~Zhang, J.~B{\"u}tepage, H.~Kjellstr{\"o}m, and S.~Mandt.
\newblock Advances in variational inference.
\newblock {\em IEEE Trans. Pattern Anal. Mach. Intell.}, 41(8):2008--2026,
  2018.

\bibitem{XZ-YL-WL:22}
X.~Zhang, Y.~Li, W.~Li, K.~Guo, and Y.~Shao.
\newblock Personalized federated learning via variational {B}ayesian inference.
\newblock In {\em Proc. Int. Conf. Mach. Learn.}, pages 26293--26310. PMLR,
  2022.

\bibitem{SZ-GYL:23}
S.~Zhou and G.~Y. Li.
\newblock {FedGiA}: An efficient hybrid algorithm for federated learning.
\newblock {\em IEEE Trans. Signal Process.}, 2023.

\end{thebibliography}
